\documentclass[10pt,twocolumn,letterpaper,dvipsnames,table,xcdraw]{article}

\usepackage[pagenumbers]{cvpr} % To force page numbers, e.g. for an arXiv version

\usepackage{bm}
\usepackage{amsmath}
\usepackage{amsfonts}
\usepackage{float}
\usepackage{tabularx, multirow}

\usepackage{array}
\newcolumntype{?}{!{\vrule width 2pt}}
\newcolumntype{Y}{>{\centering\arraybackslash}X}
\usepackage[table]{xcolor}

\definecolor{cvprblue}{rgb}{0.21,0.49,0.74}
\usepackage[pagebackref,breaklinks,colorlinks,citecolor=cvprblue]{hyperref}

\title{GASP: Gaussian Avatars with Synthetic Priors}

\author{Jack Saunders$^{1,2}$, Charlie Hewitt$^1$, 
Yanan Jian$^1$, Marek Kowalski$^1$,
Tadas Baltrusaitis$^1$, Yiye Chen$^{1,3}$, \\
Darren Cosker$^{1,2}$, Virginia Estellers$^1$,
Nicholas Gydé$^1$, Vinay P. Namboodiri$^2$,
and Benjamin E Lundell$^1$
\\
$^1$Microsoft, $^2$ University of Bath, $^3$ Georgia Tech \\
}

\begin{document}

% Title Figure
\twocolumn[{%
\renewcommand\twocolumn[1][]{#1}%
\maketitle
\vspace{-0.5cm}
\begin{center}
    \centering
    \captionsetup{type=figure}
    \includegraphics[width=\textwidth]{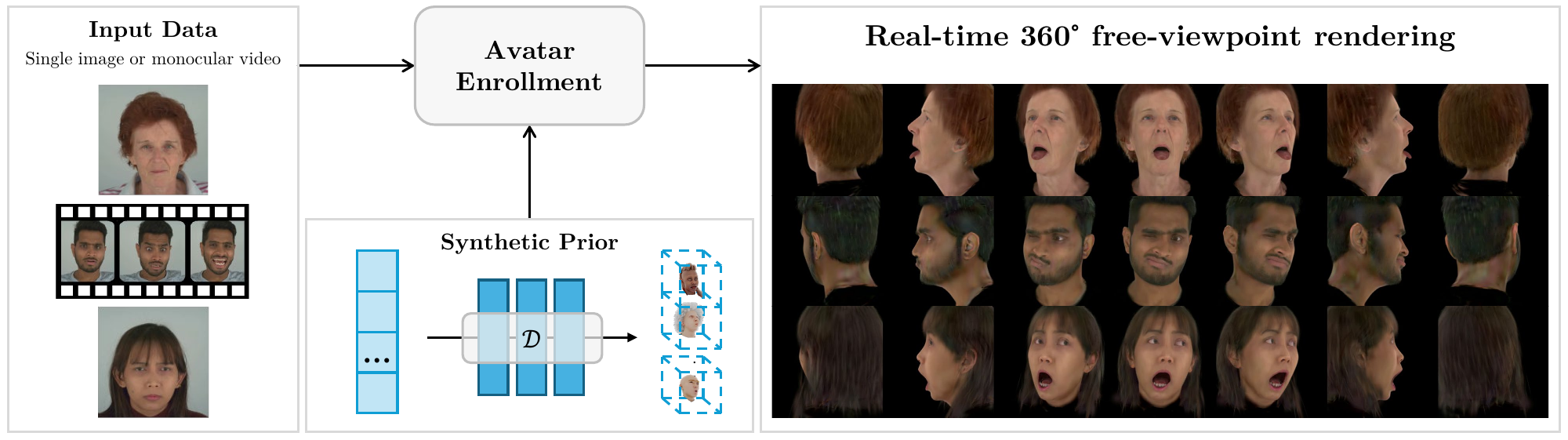}
    \caption{We propose \textbf{GASP}, a novel model for creating photorealistic, realtime, animatable, 360$^\circ$ avatars from easily-captured data. We train a generative prior model of Gaussian Avatars on Synthetic data. The prior allows our model to be fit using a single image or a short video with the prior accounting for the unseen views. This lets users create their avatar with only a webcam or smartphone.}
    \label{fig:title}
\end{center}%
}]

\maketitle

\begin{abstract}
Gaussian Splatting has changed the game for real-time photo-realistic rendering. One of the most popular applications of Gaussian Splatting is to create animatable avatars, known as Gaussian Avatars. Recent works have pushed the boundaries of quality and rendering efficiency but suffer from two main limitations. Either they require expensive multi-camera rigs to produce avatars with free-view rendering, or they can be trained with a single camera but only rendered at high quality from this fixed viewpoint. An ideal model would be trained using a short monocular video or image from available hardware, such as a webcam, and rendered from any view. To this end, we propose \textbf{GASP: Gaussian Avatars with Synthetic Priors}. To overcome the limitations of existing datasets, we exploit the pixel-perfect nature of synthetic data to train a Gaussian Avatar prior. By fitting this prior model to a single photo or video
and fine-tuning it, we get a high-quality Gaussian Avatar, which supports 360$^\circ$ rendering. Our prior is only required for fitting, not inference, enabling real-time application. Through our method, we obtain high-quality, animatable Avatars from limited data which can be animated and rendered at 70fps on commercial hardware. See our project page \footnote{https://microsoft.github.io/GASP/} for results.
\end{abstract}
\section{Introduction}
\label{sec:intro}

Creating high-quality digital humans unlocks significant potential for many applications, including Virtual Reality, gaming, video conferencing, and entertainment. Digital humans must be photorealistic, easy to capture and capable of real-time rendering. The vision and graphics communities have long worked towards this goal, and we are rapidly approaching the point where such digital humans are possible. 

A series of works based first on NeRFs \cite{Mildenhall20eccv_nerf} raised the bar in creating exceptional visual quality \cite{guo2021adnerf, Zielonka2022InstantVH, kirschstein2023nersemble, Gafni_2021_CVPR, buehler2024cafca, buhler2023preface}. However, NeRFs remain slow to render and are unsuitable for real-time applications. Gaussian Splatting-based works have led to significant improvements in both quality and rendering speed \cite{kerbl3Dgaussians, qian2023gaussianavatars, xu2023gaussianheadavatar, chen2023monogaussianavatar, shao2024splattingavatar, xiang2024flashavatar, giebenhain2024npga}. Despite these improvements, the list of suitable applications for these methods is small. Each of these models suffers from one of two drawbacks: either they require expensive capture setups with multiple synchronized cameras, which prevents easy user enrollment \cite{qian2023gaussianavatars, xu2023gaussianheadavatar, giebenhain2024npga}, or they train on a single camera but exhibit significant quality degradation when rendered from views with more than a minimal variation in camera pose \cite{xiang2024flashavatar, shao2024splattingavatar, chen2023monogaussianavatar}. Furthermore, to maximize visual quality, some of these methods use a large CNN \emph{after} rendering, which prevents real-time rendering without a powerful GPU \cite{giebenhain2024npga, xu2023gaussianheadavatar}.

For mass adoption, an avatar model should achieve high-quality 360$^{\circ}$ rendering in real-time and require only the amount of data a user can practically provide. In most cases a user can only capture a monocular, frontal image or video using their webcam or smartphone camera. The problem of fitting an avatar to this data is ill-posed; the extreme sides and back of the head are not visible, leading to artifacts in these unseen regions. In order to overcome this, we require a prior model that is able to ``fill in the gaps" left by missing data. Such a model has been shown to be effective in other limited data human-centric models, such as visual dubbing \cite{Saunders2024D4E} and static NeRF models \cite{buehler2024cafca}. Ideally, we would train such a model on a large, multi-view, perfectly annotated and diverse dataset. However, very few multi-view face datasets exist. Those that do either lack full coverage, particularly around the back of the head \cite{kirschstein2023nersemble}, or have only a small number of subjects \cite{wuu2022multiface}. Furthermore, annotations such as camera calibrations and 3D morphable model (3DMM) parameters associated with these datasets have to be estimated using imperfect methods and are a significant source of error.

% To address this problem and make progress towards an easily accessible photorealistic avatar model, 
We propose \textbf{GASP: Gaussian Avatars with Synthetic Priors}. We use a large, diverse dataset of \emph{synthetic} humans \cite{wood2021fake, hewitt2024look} to overcome the difficulties associated with training a prior on real data. This data is generated using computer graphics and has perfectly accurate annotations, including exact correspondence to the underlying 3DMM. This enables the large-scale training of a prior for Gaussian Avatars for the first time. However, the use of synthetic data introduces a domain gap problem. We address this by learning per-Gaussian features with semantic correlations. By learning these correlations on synthetic data and then maintaining them when fitting to real data, using a three-stage fitting process, we can cross this domain gap. Our method even enables rendering the back of the head, having fit to only a single front-facing image or video; see \cref{fig:title}.

To summarize, we propose a novel system for creating realistic, real-time animatable avatars from a webcam or smartphone enabled by the following contributions:

\begin{itemize}
    \item{A prior model over Gaussian Avatar parameters trained using purely synthetic data.}
    \item{A three-stage fitting process, combined with learned per-Gaussian correlations to overcome the synthetic-to-real domain gap and allow for 360$^\circ$ rendering.}
    \item{Real-time rendering enabled through use of neural networks only during training and fitting, and not at inference time.}
\end{itemize}
\section{Related Work}

\subsection{Photorealistic Animatable Avatars}

A significant number of works have attempted to build photorealistic 3D Avatars that can be animated. Most of these works use an existing animatable model, known as a 3D morphable model (3DMM) \cite{blanzvetter3DMM, FLAME:SiggraphAsia2017}. Earlier works improve the realism of a 3DMM in image space using compositing \cite{Face2Face}, a CNN model \cite{kim2018deep} or pixel-level MLPs \cite{CodecAvatars}. Some work \cite{thies2020nvp, Saunders2023READ} improve the CNN models by adding a learnable latent texture known as a neural texture \cite{DNR} and evaluating this with a deferred neural renderer. Other works make use of volumetric rendering, either in the form of a point-based representation \cite{Zheng2023pointavatar}, or a NeRF \cite{mildenhall2020nerf, guo2021adnerf, ye2023geneface, Zielonka2022InstantVH}. Each of these methods has shown great potential but is too slow or too prone to artifacts to provide a complete solution.

Gaussian Splatting \cite{kerbl3Dgaussians} has allowed for unprecedented photorealism and real-time capabilities in volumetric rendering. 
Unsurprisingly, this technology has been adapted for applications in the photorealistic avatar space. 
We refer to this class of methods as Gaussian Avatars. 
Most Gaussian Avatar methods have built upon 3DMMs as a coarse representation of the geometry and Gaussian Splatting for finer geometry and appearance.
\citet{qian2023gaussianavatars} do this by binding Gaussians individual triangles in the mesh. %, which are then transformed like the underlying triangle. 
% A binding inheritance strategy is used to spawn new Gaussians. 
\citet{xiang2024flashavatar} initialize Gaussians by sampling from a UV map and moving the Gaussians by barycentric interpolation of the posed meshes. 
They add a dynamic MLP that learns to introduce wrinkles based on the 3DMM expression blend weights.
\citet{chen2023monogaussianavatar} learn an extension to the LBS function of FLAME \cite{FLAME:SiggraphAsia2017} by extending the blendshape basis to apply per Gaussian.
\citet{xu2023gaussianheadavatar} model deformations and dynamics with MLPs and apply a CNN-based super-resolution network. 
\citet{giebenhain2024npga} do similar but attach the Gaussians to an implicit geometry model using cycle consistency. 
These methods can produce photorealistic avatars with real-time rendering and impressive quality, but either train on a single camera and evaluate on the same camera or allow novel view synthesis but only within a small range and rely on multiple cameras. 
A similar method to our work is the recent Gaussian Morphable Model of \citet{xu2023gphm}, which can fit to a single input image. 
However, as most of the training data used is from the front half of the head, it cannot produce results in the back of the head and as a new task-specific expression basis is learned, control with an existing 3DMM is impossible.

\subsection{Few-Shot Avatars}

\begin{figure*}[ht]
    \centering
    \includegraphics[width=\textwidth]{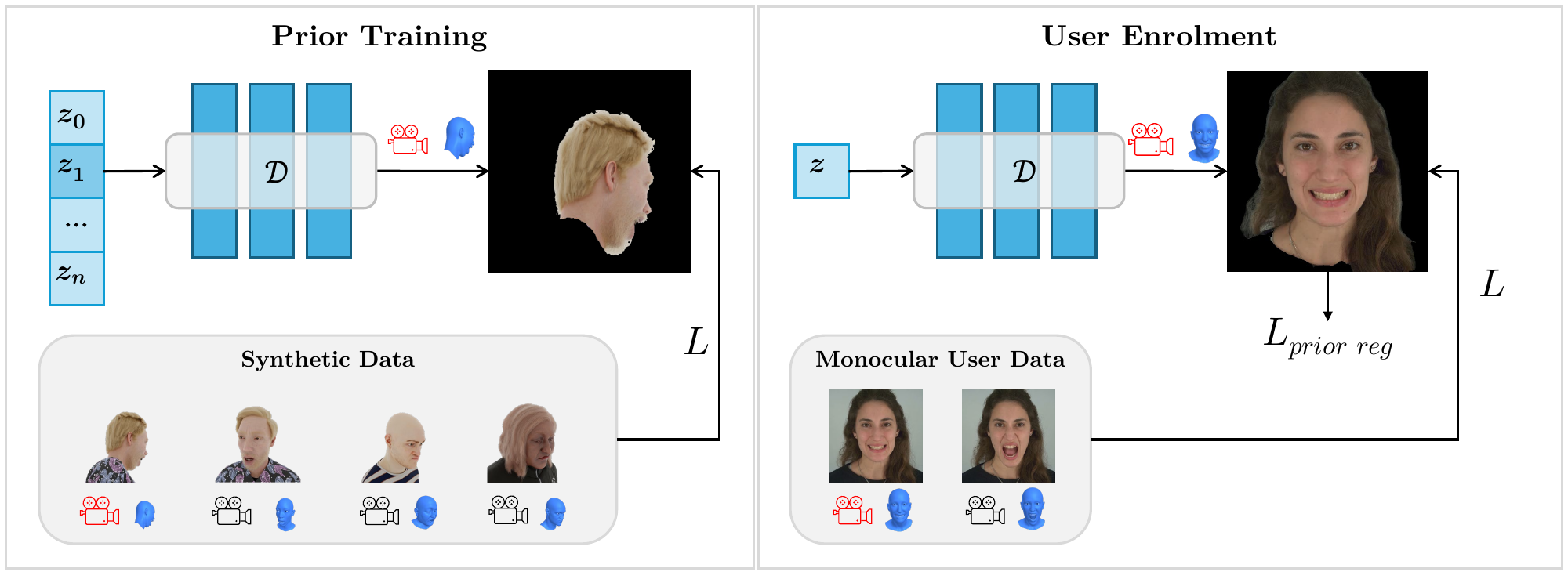}
    \caption{The overview of our model. In the first stage, we train an autodecoder prior model on Synthetic data to predict the parameters of a mesh attached Gaussian Avatar. We can then adapt this model to user enrollment data, either a single image or short monocular video. We leverage the prior to improve the quality in unseen regions and achieve free-viewpoint rendering.}
    \label{fig:arch}
\end{figure*}

Several other works have attempted to address a similar problem to ours, in which the goal is to create a photorealistic avatar from limited amounts of data. 
In each case, the solution is to leverage some form of data-driven prior. 
Preface \cite{buhler2023preface} uses a large-scale dataset to train an identity-conditioned NeRF prior model in an auto-decoder fashion.
Cafca \cite{buehler2024cafca} also seeks to train this model on large-volume synthetic data. 
While high quality, the results are static and cannot be animated, and being NeRF-based, are also slow to render, taking over 20s per frame. 
Some works use powerful 2D image-space models as a prior, exploiting a small amount of data to enable control over the larger model with a 3DMM. 
StyleRig \cite{StyleRig} first achieves control over StyleGAN2 \cite{Karras2019stylegan2} in this way, and DiffusionRig \cite{ding2023diffusionrig} obtains even better results using a DDPM \cite{ho2020denoising} as a prior. 
Dubbing for Everyone \cite{Saunders2024D4E} uses a StyleGAN-based UNET with personalized Neural Textures, which allows for better adaptation. 
ROME \cite{Khakhulin2022ROME} takes a similar approach, with neural textures predicted from images. 
% These models use a scale of data that is hard to match with 3D models.
However, as they operate at the image level, they cannot model the back and sides of the head. % as these do not appear in training data. 

%Concurrent work, HeadGAP \cite{zheng2024headgap} also builds a prior over Gaussian Avatars. This work shares a lot of similarities with ours and produces excellent results. Nonetheless, there are a few critical differences between the works. Their model requires two MLP networks and a CNN to be ran for every inference step, meaning that the model cannot be used for real-time applications. Our model only requires a neural network during training and can be run significantly faster than real-time. Furthermore, the prior trained in HeadGAP uses data of 119 real people, this limits diversity and opens up issues around privacy, as a model trained on so few people is likely to memorise these individuals and be vulnerable to dataset extraction \cite{DatasetExtraction}. In contrast, our prior model is train entirely on synthetic data, circumventing these concerns.

\section{Method}

\subsection{Background: Gaussian Splatting}

3D Gaussian Splatting is a method for reconstructing a volume from a set of images with corresponding camera calibrations. It involves using a collection of Gaussian primitives, represented by a position $\bm{\mu}$ in 3D space, an anisotropic covariance matrix $\Sigma$, a color $\mathbf{c}$ and an opacity $\bm{\alpha}$. \citet{kerbl3Dgaussians} proposed a system to optimize these parameters to fit the evidence provided by the images by decomposing the covariance $\Sigma$ into the scale, $\bm{\sigma}$, and rotation, $\mathbf{r}$, components, represented as a vector and quaternion respectively. Following projection by the camera and depth sorting, each pixel color $P$ is computed as:
\begin{equation}
    P = \sum_{i = 1}^{N_G} c_i \alpha_i \prod_{j = 1}^{i - 1} (1 - \alpha_j)
\end{equation}
Since the whole process is differentiable, the Gaussian Attributes can be optimized to match the given images and camera parameters.

\subsection{Background: Mesh Attached Gaussians}

Gaussian Splatting is excellent at reconstructing static scenes but, in its basic form, cannot model animation dynamics. 
Multiple works \cite{qian2023gaussianavatars, shao2024splattingavatar} make the observation that, given a sufficiently good coarse approximation of geometry in the form of a mesh, the problem can be reduced to an approximately static scene. 
By attaching each Gaussian, $\mathcal{G}_i$, to a specific triangle, $\mathbf{t}$, in the mesh, the Gaussian is assumed to remain static relative to that triangle's pose. There are several successful formulations of this posing transformation:

\begin{equation}
    \bm{\mu}, \bm{\sigma}, \mathbf{r} =  \mathcal{T}_{\textit{local} \rightarrow \textit{global}}(\bm{\mu}', \bm{\sigma}', \mathbf{r}' \; | \; \mathbf{t})
\end{equation}
For our purposes, we use the definition of \citet{qian2023gaussianavatars}, where the origin of each triangle's system is assumed to be the center, the orthonormal basis is determined by one edge, the triangle's normal and their cross-product, and the isotropic scale by the mean of the length of one edge and its perpendicular in the triangle. 
This allows us to define a Gaussian Avatar, $\mathcal{G}$, as a collection of static Gaussian primitives in a triangle-local space. 
\begin{equation}
    \begin{split}
    \mathcal{G} &= \{\mathcal{G}_i: 1 \leq i \leq N_G\}, \: \mathcal{G}_i = \{\bm{\mu}'_i, \bm{\sigma}'_i, \mathbf{r}'_i, \mathbf{c}_i, \mathbf{o}_i \}
    \end{split}
\end{equation}
As these are static, we can optimize them using the same procedures as in the original formulation \cite{kerbl3Dgaussians}.

\subsection{Prior Model Training}
\label{sec:prior}

We train our prior model as a generative model over identities. Following previous work \cite{Park_2019_CVPR, giebenhain2023nphm, xu2023gphm, Rao_2022_BMVC}, we train this prior as an auto-\emph{de}coder model. We jointly learn a per-subject identity code, $\mathbf{z}_j \in \mathbb{R}^{512}, j \in \{1, \ldots, N_{\textit{id}}\}$, and an MLP decoder, $\mathcal{D}(\mathbf{z})$. One may naively think of training this model to directly output the Gaussian Attributes, $\mathcal{A}$, with a single MLP. However, such a method quickly becomes intractable. As a typical model with 100,000 Gaussians may have millions of attributes, the number of parameters in $\mathcal{D}$ would be too large. Instead, we augment each Gaussian with a learnable feature vector, $\mathbf{f}_i \in \mathbb{R}^8, i \in \{1, \ldots,  N_{\textit{G}}\}$. This feature is analogous to a positional encoding with additional semantic meaning. We then train a network to map these per-Gaussian features to Gaussian attributes, with each Gaussian processed independently and in parallel.

To make optimization more stable, we learn a Canonical Gaussian Template, $\mathcal{C}$, and model the per-person variation as offsets from this template. The Canonical Template can be considered the mean Avatar. The $i$-th Gaussian of the avatar for the subject $j$ is given by:
\begin{equation}
\begin{split}
    \mathcal{A}_{i, j} = \mathcal{C}_{i, j} + \mathcal{D}(\mathbf{f}_i, \mathbf{z}_j) 
\end{split}
\end{equation}
This is best understood by following \cref{fig:stages}. To train this model, we jointly optimize $\mathcal{C}$, $\mathcal{D}$, $\{\mathbf{z}_j\}_{1 \leq j \leq N_{\textit{id}}}$, $\{\mathbf{f}_i\}_{1 \leq i \leq N_{\textit{G}}}$ to minimize the following loss function:
\begin{equation}
    \mathcal{L} = \lambda_{\textit{pix}}L_{\textit{pix}} + \lambda_{\alpha}L_{\alpha} + \lambda_{\textit{percep}}L_{\textit{percep}} + L_{\textit{reg}}
\end{equation}
Where $L_{\textit{pix}}$ is a pixel level loss consisting of $L_1$, the $\ell_1$ difference between the real and predicted images, and $L_{\textit{SSIM}}$ which is the differentiable SSIM loss, weighted by $\lambda_1$ and $\lambda_{\textit{SSIM}}$ respectively. 
$L_{\textit{percep}}$ is a perceptual loss based on LPIPS \cite{zhang2018perceptual}, $L_{\alpha}$ is the $\ell_1$ distance between the real and predicted alpha masks, and $L_{\textit{reg}}$ is a regularization loss acting on the Gaussians. 
We regularize scale and displacement:
\begin{equation}
    L_{\textit{reg}} = \lambda_{\sigma} ||max(0.6, \sigma')||_2 
    + \lambda_{\mu} ||\mu'||_2
\end{equation}
Unlike previous methods, our 3DMM does not capture course hair, meaning the Gaussians must model it. 
We, therefore, reduce $\lambda_\mu$ by a factor of 100 for Gaussians bound to faces in the scalp region, which we manually define.

\subsection{Initialization}
\label{sec:init}

Using just one Gaussian per triangle face of the 3DMM leads to an under-parameterised model that lacks sufficient detail. To overcome this, we use the initialization strategy of \citet{xiang2024flashavatar}. We generate a UV map of a given resolution for our mesh and take each pixel's corresponding face and barycentric coordinates. The face is used for Gaussian binding. We use the barycentric coordinates to position the origin of each Gaussian's local coordinate system.

%\subsection{MLP Coordinate System}

%Our decoder network, $\mathcal{D}$, predicts changes in the canonical Gaussians' positions, scales and rotations. 
%As we represent the canonical Gaussians in a triangle local coordinate frame, each Gaussian's coordinate system is different, and $\mathcal{D}$ would have to operate in many systems. 
%To overcome this, we introduce the inverse of the posing transform:
%\begin{equation}
%    \mathcal{T}_{\textit{global} \rightarrow \textit{local}} := \mathcal{T}_{\textit{local} \rightarrow \textit{global}}^{-1}
%\end{equation}
%The posing transform is a composition of simple geometric transforms based on the mesh triangles, so it is trivial to invert. 
%We can then have the MLP output attributes in the canonical space and transform them to triangle-local space with $\mathcal{T}_{\textit{global} \rightarrow \textit{local}}$ using triangles of the mean template mesh. This transform means the MLP outputs attributes in a shared canonical space and improves the quality.

\subsection{Fitting Process}

\begin{figure}
    \centering
    \includegraphics[width=\columnwidth]{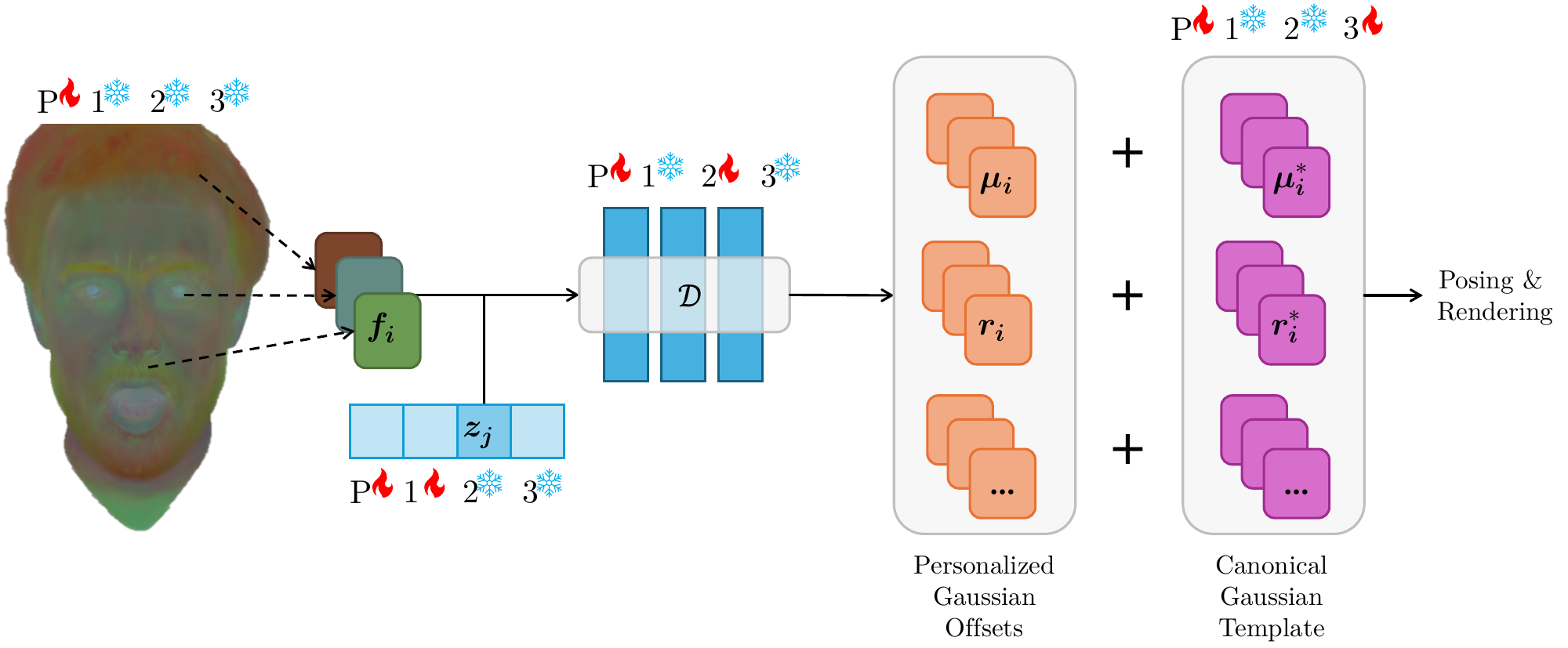}
    \caption{The architecture of our prior model. A latent vector for identity is used to transform learnable per-Gaussian features into Gaussian Attributes, which offset a canonical template. Our training process has four stages: the prior training, P, and three user-specific fitting steps. We freeze some layers and train others at each stage, as indicated.}
    \label{fig:stages}
\end{figure}

% Our prior model can improve the avatars created from a user's training data by ``filling in" missing regions in the data. 
Given input data ranging from a single image, to a short video from a single monocular camera, we aim to produce a high-quality avatar that can be viewed from any angle. %We refer to this as the fitting process. 
We have three stages to this fitting process, visualized in \cref{fig:stages}:

\begin{enumerate}
    \item We find the best in-prior Gaussian Avatar by randomly initializing an identity latent vector, $\mathbf{z}$, and optimizing this with everything else frozen; we call this inversion.  
    \item We fine-tune the MLP, $\mathcal{D}$, with the rest of the model frozen.
    \item We refine the resulting Gaussians using the standard Gaussian Splatting optimization procedure \cite{kerbl3Dgaussians} to best fit the data.
\end{enumerate}

To motivate this three-step process, we can consider two extremes. 
On the one hand, we could perform inversion only. This relies heavily on the prior. 
If we had perfectly diverse real-human data and a perfect prior, this may be all we would need to do. 
However, our prior training was on synthetic data, so we could only generate synthetic-looking avatars with this method. 
On the other hand, we could use the prior for initialization and then optimize the resulting Gaussians. 
This would achieve similar results to the existing state-of-the-art but with the unseen regions looking synthetic. 

\begin{figure}
    \centering
    \includegraphics[width=\columnwidth]{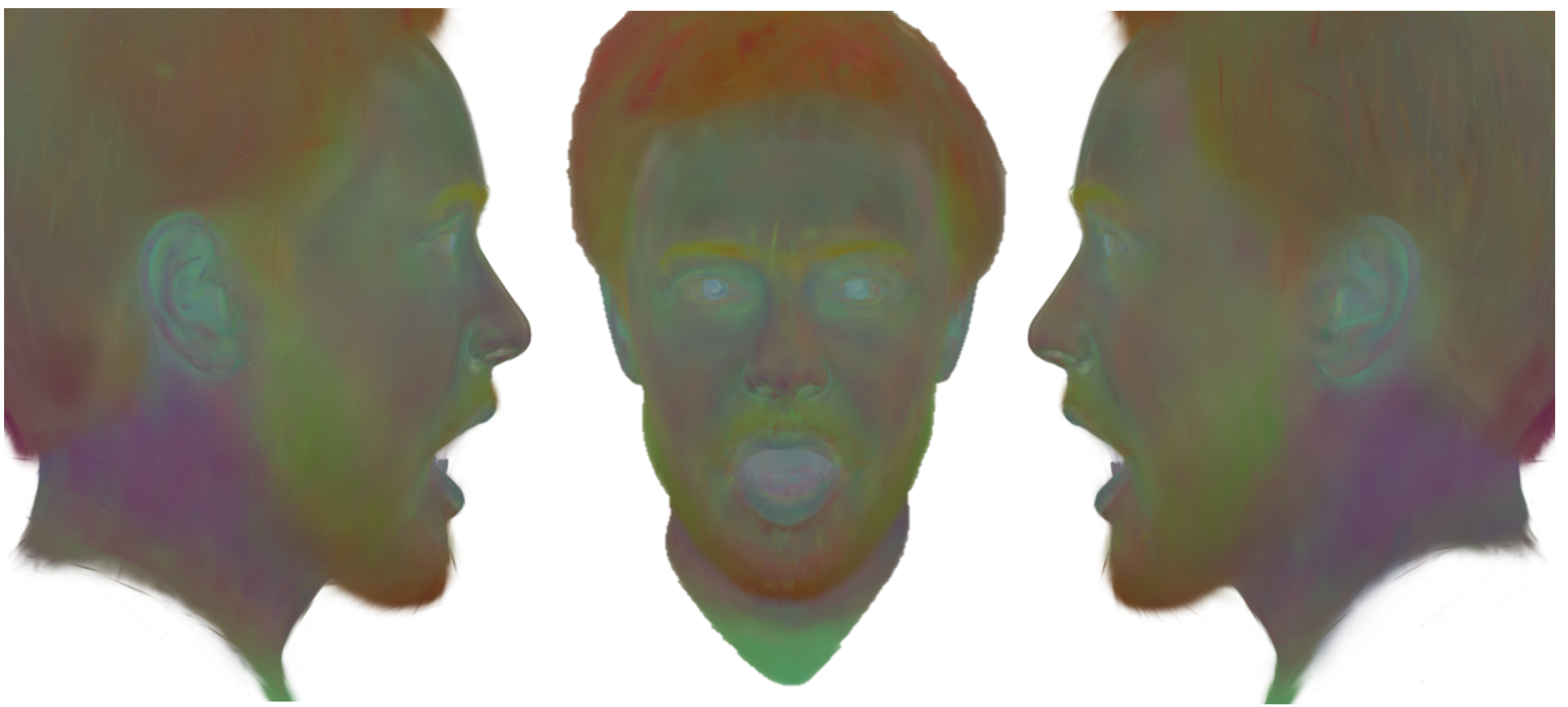}
    \caption{Visualization of the first three components of a PCA decomposition of the Gaussian features $\mathbf{f}$, displayed using the geometry of a random subject. Note the semantic relationships.}
    \label{fig:PCA}
\end{figure}

We can extract more value from our prior model by considering correlations in the per-Gaussian features, $\mathbf{f}$. 
Our network is forced to map these to Gaussian attributes and learns to associate similar Gaussians with similar features. 
\cref{fig:PCA} shows a PCA decomposition of the Gaussian features, demonstrating that these features have learned semantic meaning. 
By freezing the features in the fitting process, Gaussians with similar semantic features will be mapped to have similar attributes. 
For example, if $\mathcal{D}$ learns to make a Gaussian representing hair at the front of the head blonde, it will also update an unseen one at the back of the head.

\begin{figure}
    \centering
    \includegraphics[width=0.9\columnwidth]{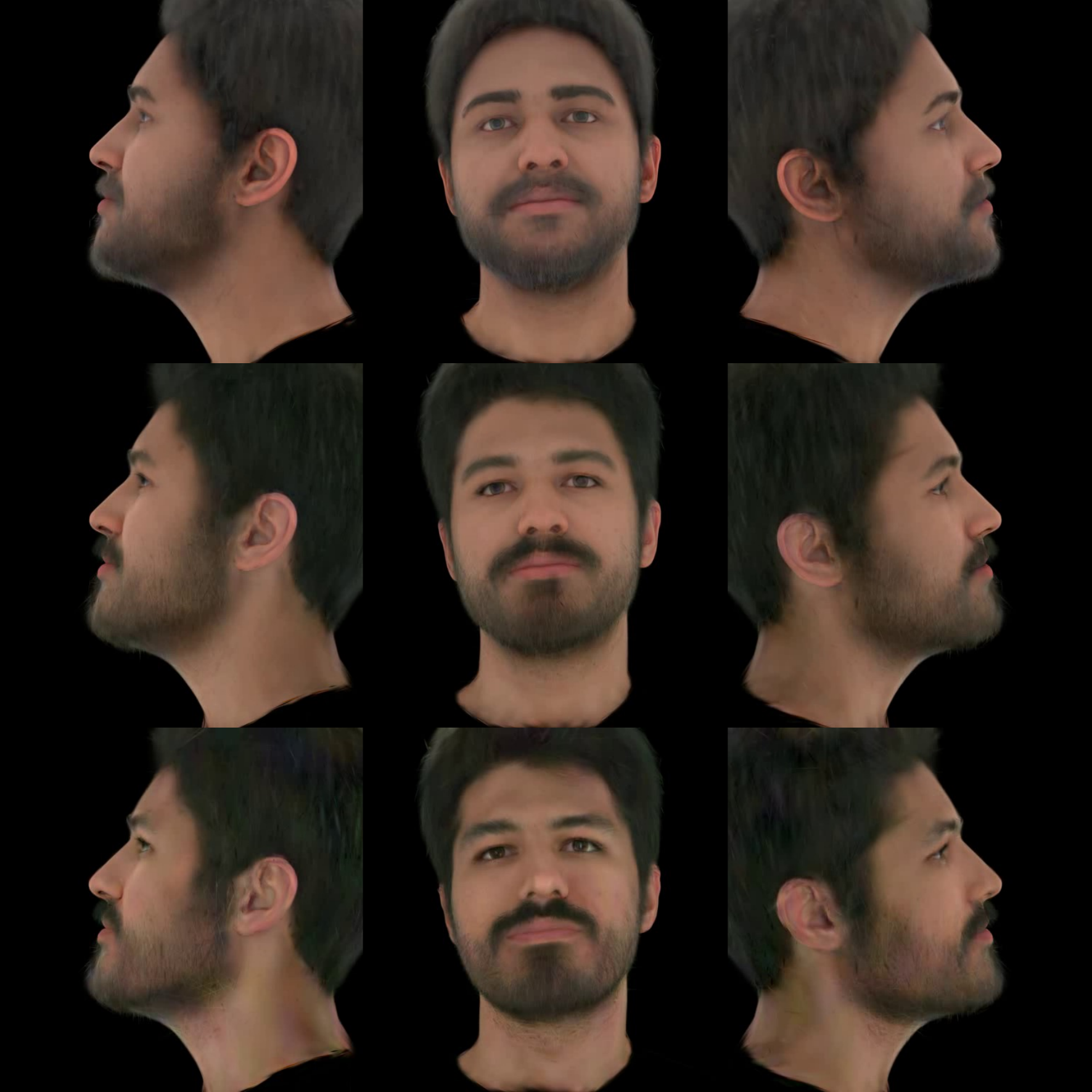}
    \caption{Examples showing how the three stages in our fitting process resolve the domain gap of the synthetic prior. Stage 1 (Top) optimizes within the prior, Stage 2 (Middle) finetunes the MLP, $\mathcal{D}$, and Stage 3 (Bottom) refines the individual Gaussians. Note the beard and eyes.}
    \label{fig:stages_qual}
\end{figure}

% \paragraph{Prior Regularization:}
To prevent stages 2 and 3 from diverging too far from the prior, we introduce an additional regularization term, $L_{\textit{prior}}$, to the loss, $\mathcal{L}$, during these stages.
% regularize the fitting process during these stages. 
% In addition to the loss $\mathcal{L}$ used in prior training, we include a prior regularization term $L_{\textit{prior}}$. 
$L_{\textit{prior}}$ is defined as the $\ell_2$ distance between each Gaussian attribute and its corresponding value from the prior (i.e., after stage 1). 
This is particularly important when regularizing unseen Gaussians.
Results after each stage of fitting are shown in \cref{fig:stages_qual}.

\section{Dataset}

\begin{table*}[t]
\centering
\resizebox{\textwidth}{!}{%
\begin{tabular}{l|cccccc|cccccc}
\toprule
 & \multicolumn{6}{c}{Monocular Video} & \multicolumn{6}{c}{Single Image} \\
\midrule
Method & PSNR $\uparrow$ & SSIM $\uparrow$ & LPIPS $\downarrow$ & FID $\downarrow$ & ID-SIM $\uparrow$ & QUAL $\uparrow$ & PSNR $\uparrow$ & SSIM $\uparrow$ & LPIPS $\downarrow$ & FID $\downarrow$ & ID-SIM $\uparrow$ & QUAL $\uparrow$ \\
\midrule
FlashAvatar & 17.25 & \cellcolor{yellow!25}0.603 & 0.450 & 351 & 0.234 & 2.08 & 13.26 & 0.490 & 0.519 & 367 & 0.057 & 2.05 \\
GaussianAvatars & 17.39 & 0.601 & \cellcolor{yellow!25}0.428 & 366 & 0.179 & 2.08 & 14.80 & 0.474 & 0.475 & 385 & 0.000 & 2.03 \\
ROME* & - & - & - & - & - & - & 15.78 & \cellcolor{yellow!25}0.543 & \cellcolor{yellow!25}0.441 & \cellcolor{yellow!25}136 & \cellcolor{yellow!25}0.408 & \cellcolor{yellow!25}3.38 \\
DiffusionRig & \cellcolor{yellow!25}19.67 & 0.343 & 0.436 & \cellcolor{yellow!25}155 & \cellcolor{yellow!25}0.302 & \cellcolor{yellow!25}2.98 & \cellcolor{yellow!25}16.87 & 0.316 & 0.541 & 183 & 0.239 & 3.15 \\
Ours & \cellcolor{red!25}21.34 & \cellcolor{red!25}0.712 & \cellcolor{red!25}0.333 & \cellcolor{red!25}117 & \cellcolor{red!25}0.568 & \cellcolor{red!25}3.68 & \cellcolor{red!25}20.73 & \cellcolor{red!25}0.677 & \cellcolor{red!25}0.348 & \cellcolor{red!25}119 & \cellcolor{red!25}0.526 & \cellcolor{red!25}3.80 \\
\bottomrule
\end{tabular}
}
\caption{
\textbf{Quantitative Evaluations: } We compare our method with three state-of-the-art models. We evaluate on two scenarios, for the Monocular scenario we on a single camera and then evaluate on the four most extreme. For single image we do the same but using only the first image from the Monocular sequence. In each case the evaluation sequence is unseen in the training set. We take the average PSNR, SSIM and LPIPS scores for each frame of each avatar. We also ask for user ratings of the quality of each method and report the mean scores out of 5 (QUAL). (*) ROME only supports single image use cases. We highlight the \colorbox{red!25}{Best} and \colorbox{yellow!25}{Second Best} for each metric.
}
\vspace{-0.05in}
\label{tab:main_quant}
\end{table*}

\begin{figure}
    \centering
    \includegraphics[width=\columnwidth]{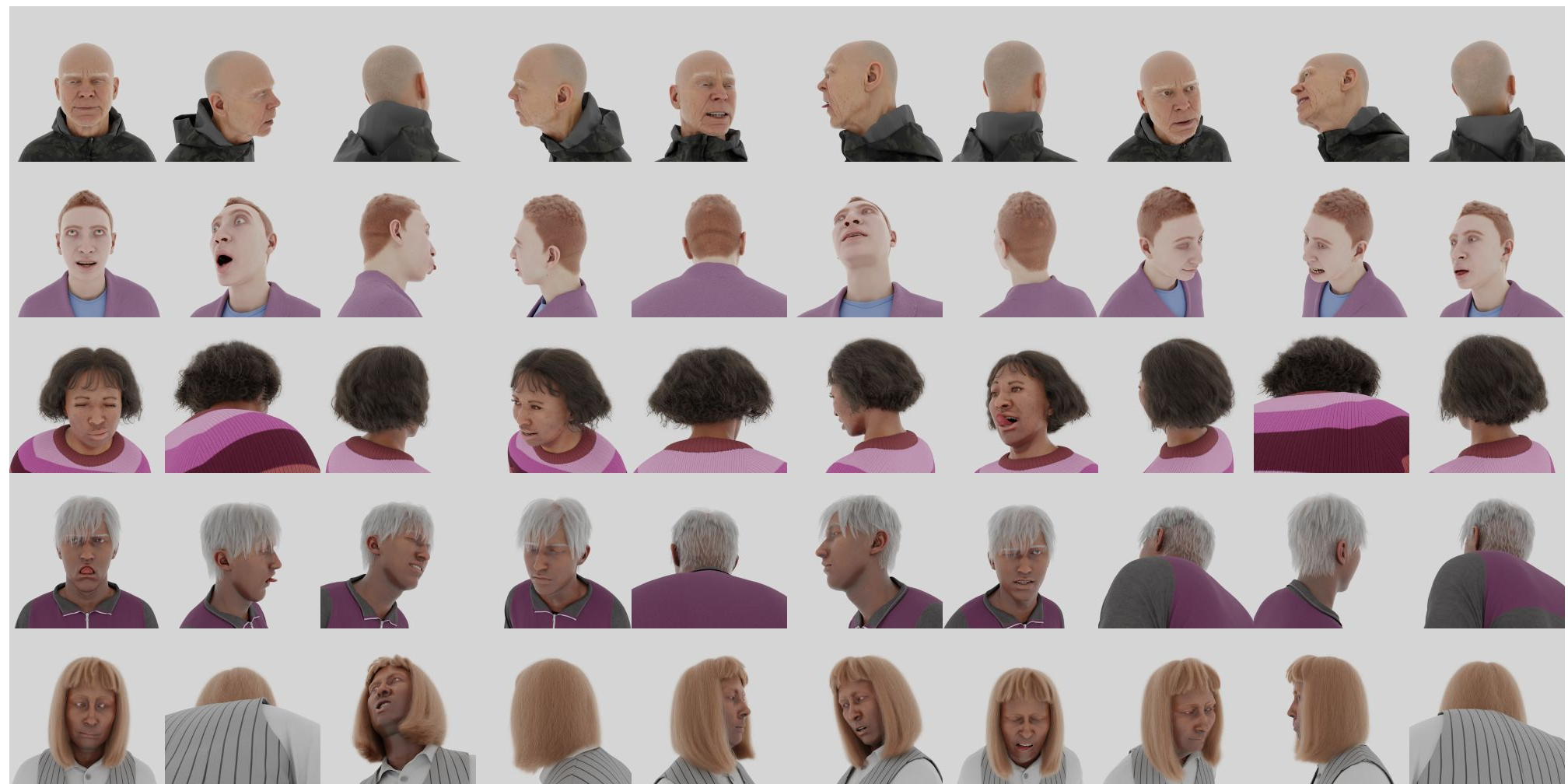}
    \caption{Examples from our synthetic dataset. We generate a large and diverse set of synthetic subjects rendered from many views to train our prior model.}
    \label{fig:synthetics}
\end{figure}

We require calibrated multi-camera data of the same subject performing a wide range of expressions to train our prior model.
Collecting such data would require complex and expensive camera rigs. 
Instead, we leverage a synthetic data generation pipeline of \citet{hewitt2024look}.
This allows us to generate highly diverse and perfectly calibrated image data with pixel-perfect annotations.
We generate 1000 identities (random face shape, texture, upper body clothing, hairstyle, hair color, and eye color).
We illuminate the scene using uniform white lighting to simplify model training.
We pose those faces with random expressions and sample a virtual camera uniformly from a hemisphere ([-180, +180] degrees azimuth and [-20, +45] degrees elevation) to render 50 images per identity.
Examples of the data used in training our prior is shown in \cref{fig:synthetics}.

%Ideally, we would train our prior model with large quantities of real-world, multi-camera data. However, this is challenging to obtain as it requires recording thousands of subjects in expensive studios. Instead, we use the synthetic data generation pipeline of \citet{wood2021fake}; this allows us to generate pixel-perfect data with high levels of diversity. To simplify the problem and improve prior model training, we reduce the additional degrees of freedom in the data by using only uniform, white lighting. We also remove accessories from this version of the dataset. We generate 1000 identities by sampling identity blendshape coefficients, hairstyles and textures. We sample 30 images for each identity, each with random expression and camera pose. Unlike previous work, we include samples of the back of the head. 
\section{Implementation Details}

Identity codes, $\mathbf{z}$, and Gaussian features, $\mathbf{f}$, are 512 and 8 dimensional, respectively. 
We initialize with a UV map of $512\times512$ pixels, resulting in 187,779 Gaussians. 
The supplementary details our decoder network $\mathcal{D}$'s architecture.
For all parameters, we optimize using the Adam optimizer \cite{Kingma2014AdamAM}. 
The canonical Gaussians are optimized using the learning rates from the original implementation of 3D Gaussian Splatting \cite{kerbl3Dgaussians}. 
We optimize $\mathbf{z}$ and $\mathcal{D}$ with a learning rate 0.0002. 
Prior network training took 4 days and was performed using 4$times$A100's with a batch size of 8 for 250 epochs. 
The fitting process uses 500 steps for stages 1 and 2. 
We use 100 steps for stage 3. The whole fitting process takes 10 minutes on an NVIDIA Geforce 4090 RTX GPU.
\section{Results}
\begin{table}[t]
\centering
\resizebox{\columnwidth}{!}{%
\begin{tabular}{l|cccccc}
\toprule
Method & PSNR $\uparrow$ & SSIM $\uparrow$ & LPIPS $\downarrow$ & FID $\downarrow$ & ID-SIM $\uparrow$ & QUAL $\uparrow$ \\
\midrule
FlashAvatar & \cellcolor{red!25}24.73 & \cellcolor{red!25}0.815 & \cellcolor{red!25}0.253 & 125 & \cellcolor{yellow!25}0.767 & \cellcolor{yellow!25}3.70 \\
GaussianAvatars & \cellcolor{yellow!25}23.73 & \cellcolor{yellow!25}0.812 & 0.285 & \cellcolor{yellow!25}113 & \cellcolor{red!25}0.773 & 3.65  \\
DiffusionRig & 19.42 & 0.377 & 0.425 & 155 & 0.302 & 3.00 \\
Ours & 23.44 & 0.786 & \cellcolor{yellow!25}0.261 & \cellcolor{red!25}101 & 0.734 & \cellcolor{red!25}3.80 \\
\bottomrule
\end{tabular}
}
\caption{
\textbf{Multi-Camera: } We run comparisons using 16 training cameras. 
%We take the average PSNR, SSIM and LPIPS scores for each frame of each avatar. 
We report the mean user ratings out of 5 (QUAL). We highlight the \colorbox{red!25}{Best} and \colorbox{yellow!25}{Second Best} for each metric.
}
\label{tab:multicam_quant}
\end{table}

\begin{figure*}
    \centering
    \footnotesize
    \begin{tabularx}{\textwidth}{@{}YYYYYY@{}}
         \textbf{Diffusion Rig} & \textbf{Flash Avatars} & \textbf{GaussianAvatars} & \textbf{Ours} & \textbf{Ours} & \textbf{Ground} \\
         \cite{ding2023diffusionrig} & \cite{xiang2024flashavatar} & \cite{qian2023gaussianavatars} & \textbf{(no prior)} & \textbf{Full} & \textbf{Truth} \\
    \end{tabularx}
    \includegraphics[width=\textwidth]{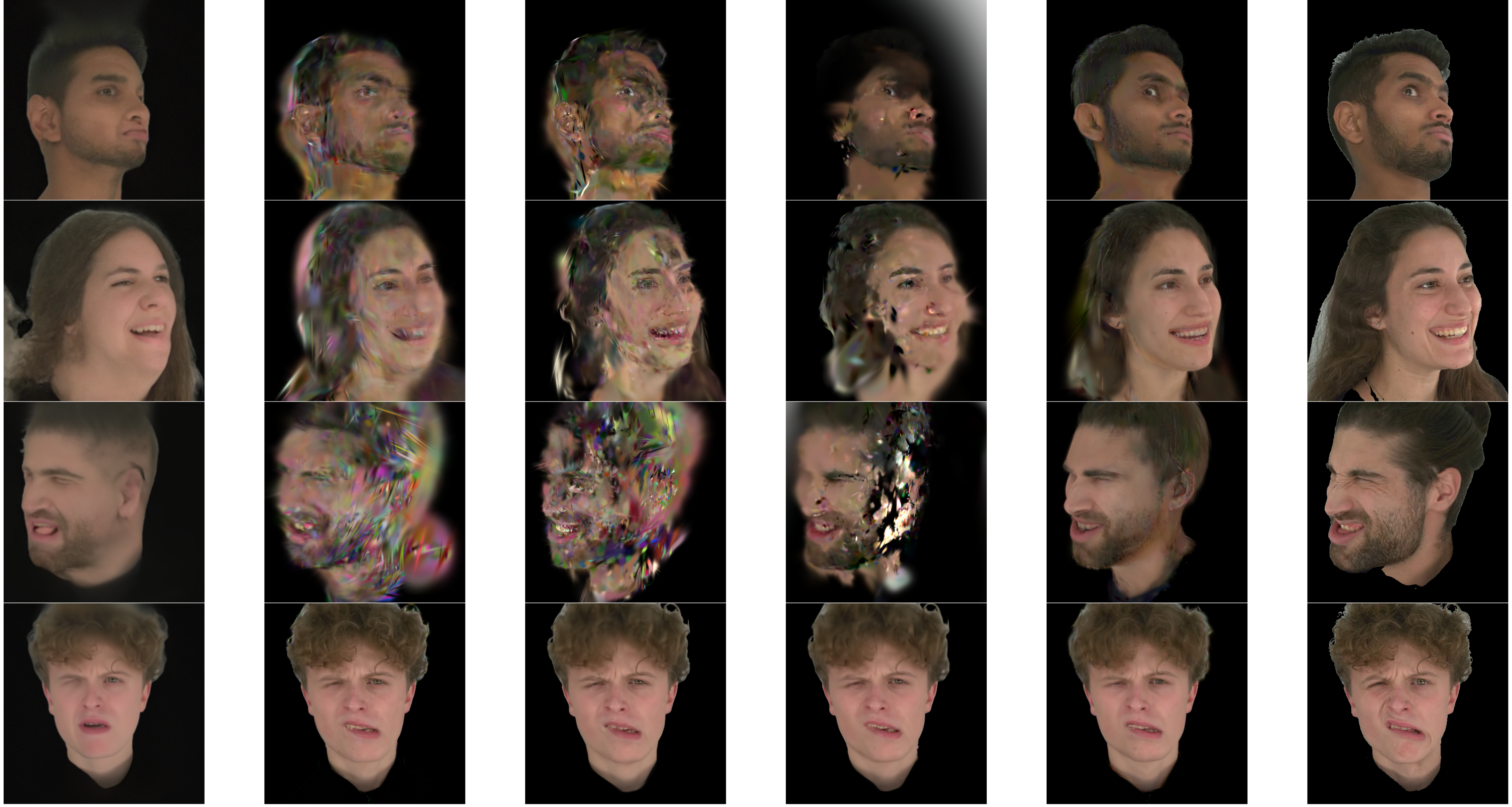}
    \caption{Qualitative comparisons of our method with existing state-of-the-art in the \textbf{Monocular Setting}. We train on a monocular camera and evaluate on unseen camera poses (top three rows) and an unseen sequence from the training view (bottom row). Our model captures identity better than Diffusion Rig \cite{ding2023diffusionrig} and suffers from fewer artifacts than other Gaussian Avatar models (\cite{qian2023gaussianavatars, xiang2024flashavatar}, ours without a prior)}
    \label{fig:main_qual}
\end{figure*}

\begin{figure*}[!ht]
    \centering
    \includegraphics[width=0.9\textwidth]{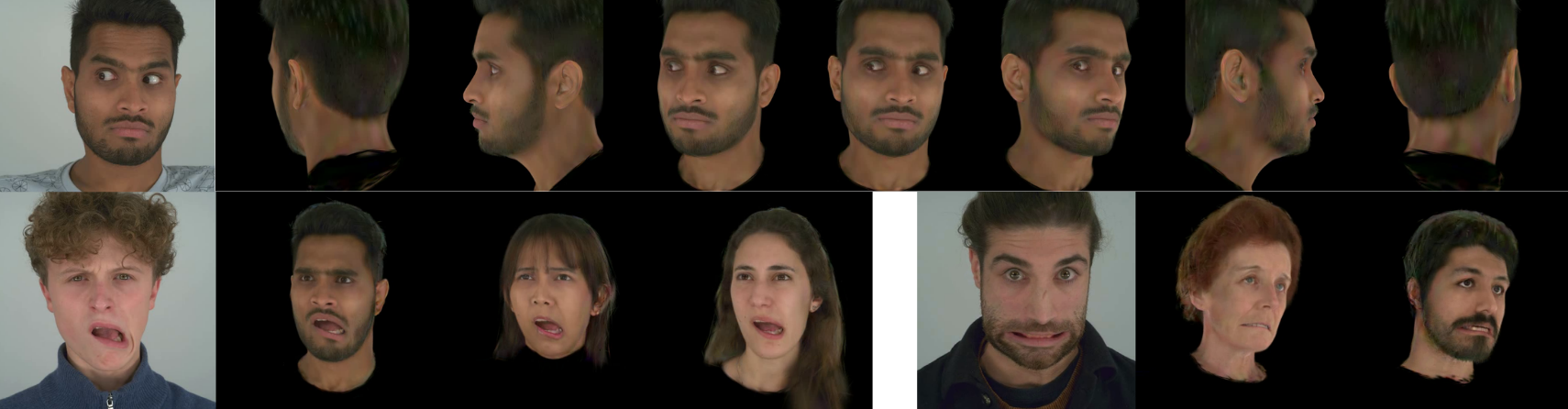}
    \caption{\textbf{Self/Cross Reenactment: } We show examples of our model for self-reenactment (top) and cross-identity (bottom). The model is fit using a frontal view video only (frame with a gray background). Despite never seeing the back of a real person's head, we still obtain good-quality results (frames with a black background). More examples are in \cref{fig:title} and the supplementary.}
    \label{fig:self-and-cross}
\end{figure*}

We conduct all of our evaluations on the NeRSemble Dataset \cite{kirschstein2023nersemble}. This dataset contains multiple subjects performing dozens of facial expression sequences, including one freeform sequence, across 16 cameras. For each sequence, we preprocess each video using an off-the-shelf background removal \cite{rvm} and face segmentation tool \cite{BiSeNet} to get the head region only. We obtain Morphable Model parameters in the format of \citet{wood2021fake} using the method of \citet{hewitt2024look}.
We consider three experimental settings using this data; please refer to the supplementary material for the cameras and sequences used: 

\emph{Monocular:} To best replicate our desired setting, we enroll all avatars using a single frontal camera. 
We use a subset of the expression sequences for fitting and evaluate them using the unseen freeform sequence. 
We use the four most extreme view cameras for evaluation, as determined by manual inspection, to test the model's ability to produce good results on regions unseen at training. 

\emph{Multi-Camera:} To confirm that our model does not sacrifice performance when more data is available, we also enroll avatars using the same configuration above but with all cameras used for input.
% the same subset of expression sequences but using the same unseen freeform sequence with the same extreme cameras for evaluation. 

\emph{Single Image:} To test the limits of our model, we also experiment with just a single image as input, selecting the first frame from the Monocular setting as input.

To evaluate visual quality, we use the standard metrics PSNR, SSIM, LPIPS \cite{zhang2018perceptual} and FID \cite{FID17}. We find that PSNR and SSIM prefer solutions that match low-frequency detail, e.g. a flat sheet of hair. While FID is better at capturing high-frequencies. We also conducted a user study to measure perceived quality most accurately. We ask users to rate each video out of five and report the mean scores; we denote this QUAL. More details can be found in the supplementary. 

\subsection{Baselines}

We compare our model to state-of-the-art methods. For the first set of methods, we look at Gaussian Avatar models: Gaussian Avatars \cite{qian2023gaussianavatars}, which is designed for ultra-high quality rendering when trained on multiple views, and Flash Avatars \cite{xiang2024flashavatar}, which is designed to be trained and evaluated on monocular data. We train these using the same morphable model, 3DMM fitting process and dataset preprocessing as our method. In addition to Gaussian Avatar models, we look at models designed for few-shot animatable avatar synthesis. We select the publicly available implementations of ROME \cite{Khakhulin2022ROME} and DiffusionRig \cite{ding2023diffusionrig}.

\subsection{Monocular Training}

The results of this experiment can be found in \Cref{tab:main_quant}. Our model significantly outperforms state-of-the-art across all metrics, including user-perceived quality. Our model produces significantly fewer artifacts in novel views compared to other Gaussian Avatar methods \cite{xiang2024flashavatar, qian2023gaussianavatars}. This is because our prior helps prevent the model from overfitting to the training camera view. Diffusion Rig \cite{ding2023diffusionrig} does not show any visible artifacts, but struggles to preserve the identity of the subject, this is best seen in \cref{fig:main_qual}.

\subsection{Multi-Camera Training}

Our model is competitive with the state-of-the-art in the Multi-Camera setting (\cref{tab:multicam_quant}). We expect our model to perform worse than other Gaussian Avatar methods \cite{qian2023gaussianavatars, xiang2024flashavatar} as the prior regularizes the model towards a synthetic solution, and we do not model lighting or dynamic expressions. Despite this, our model performs similarly to the state-of-the-art, suggesting it can effectively use all available data. Furthermore, using the prior allows our model to converge in fewer steps than other Gaussian Avatar models, making it cheaper and more efficient to train. Our model performs better on all metrics compared to Diffusion Rig \cite{ding2023diffusionrig}. 

\subsection{Single Image Training}

%\begin{figure}
%    \centering
%    \includegraphics[width=\columnwidth]{media/GASP Single Image Resulst.png}
%    \caption{Qualitative comparisons of our method with existing state-of-the-art in the \textbf{Single Image Setting}, using the top image.}
%    \label{fig:main_qual}
%\end{figure}

\begin{figure}
    \centering
        \footnotesize
        \begin{tabularx}{\linewidth}{ll}
        \centering
            \begin{tabular}{l}
                \textbf{Input} \\ \textbf{Image}
            \end{tabular}& \raisebox{-.5\height}{\includegraphics[width=0.6\linewidth]{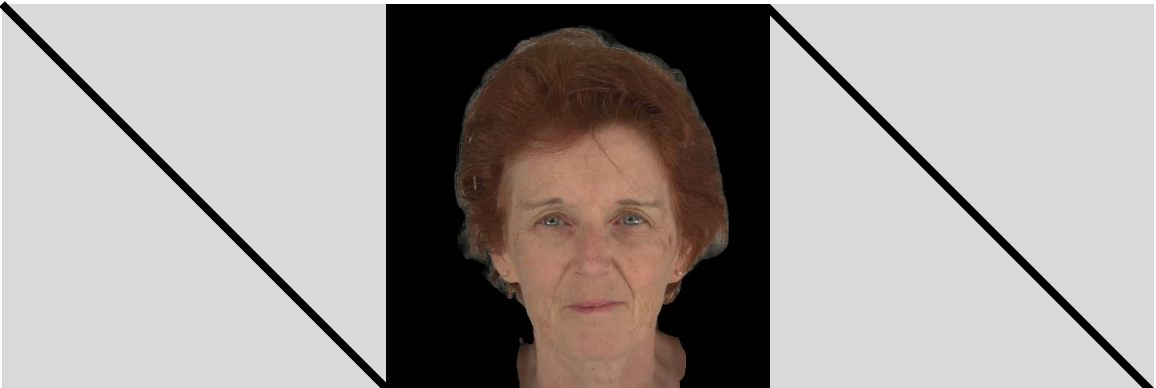}} \\
            \begin{tabular}{l}
                 \textbf{ROME} \cite{Khakhulin2022ROME}
            \end{tabular}  & \raisebox{-.5\height}{\includegraphics[width=0.6\linewidth]{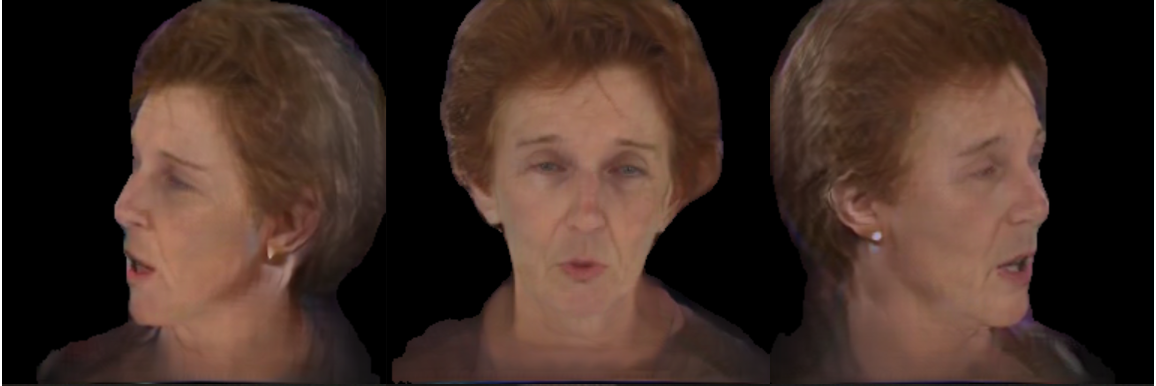}} \\
            \begin{tabular}{l}
                 \textbf{Diffusion} \\ \textbf{Rig} \cite{ding2023diffusionrig}
            \end{tabular} & \raisebox{-.5\height}{\includegraphics[width=0.6\linewidth]{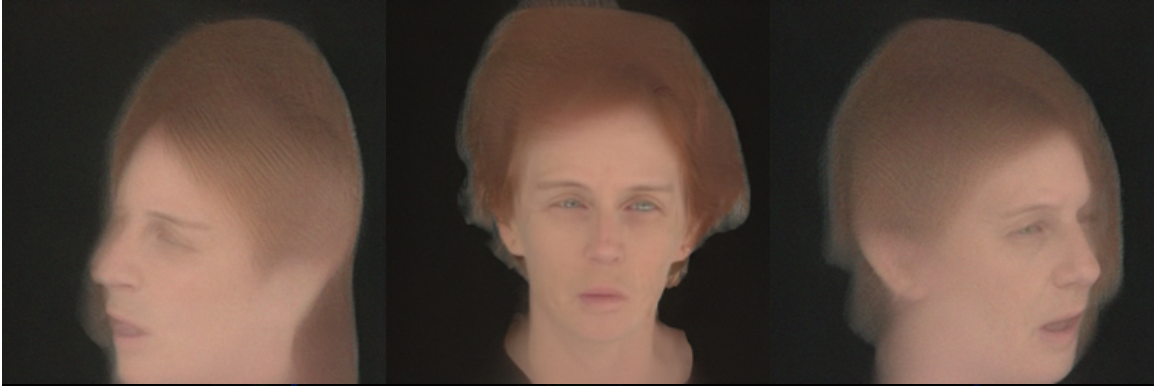}} \\
            \begin{tabular}{l}
                 \textbf{Gaussian} \\ \textbf{Avatars} \cite{qian2023gaussianavatars}
            \end{tabular} & \raisebox{-.5\height}{\includegraphics[width=0.6\linewidth]{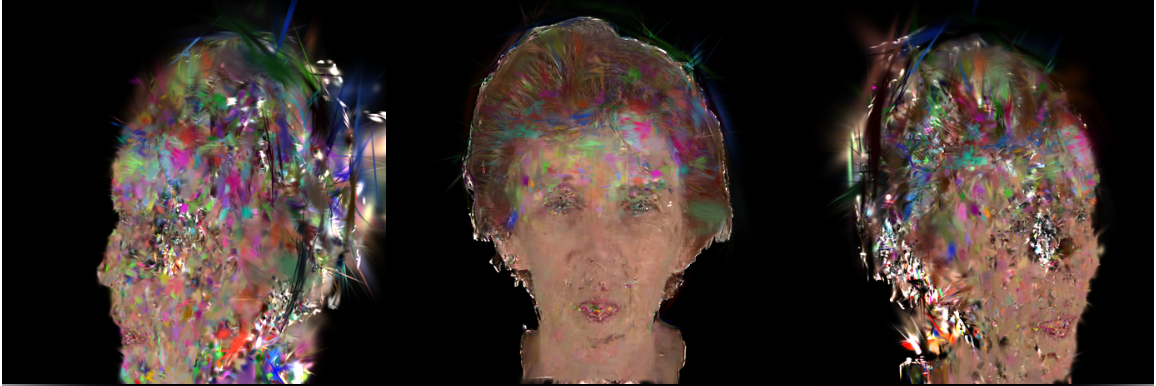}} \\
            \begin{tabular}{l}
                 \textbf{Ours} \\ \textbf{(No Prior)}
            \end{tabular}  & \raisebox{-.5\height}{\includegraphics[width=0.6\linewidth]{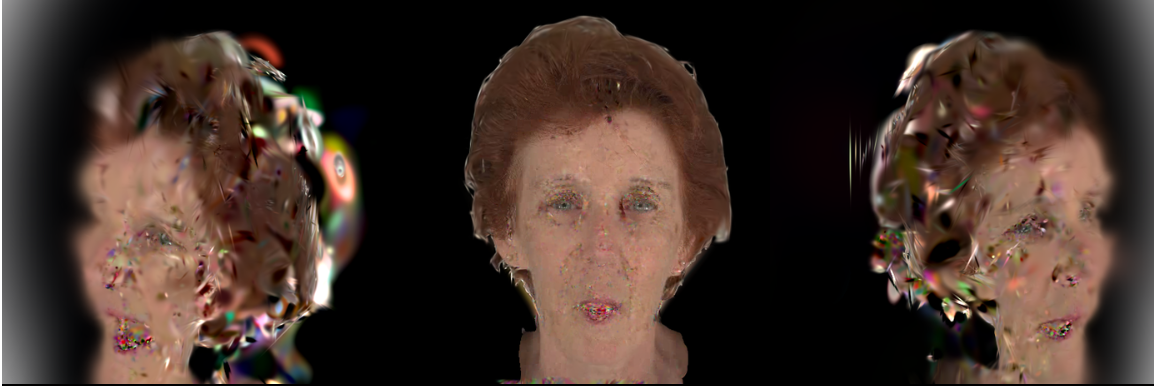}} \\
            \begin{tabular}{l}
                 \textbf{Ours} \\ \textbf{Full} 
            \end{tabular}  & \raisebox{-.5\height}{\includegraphics[width=0.6\linewidth]{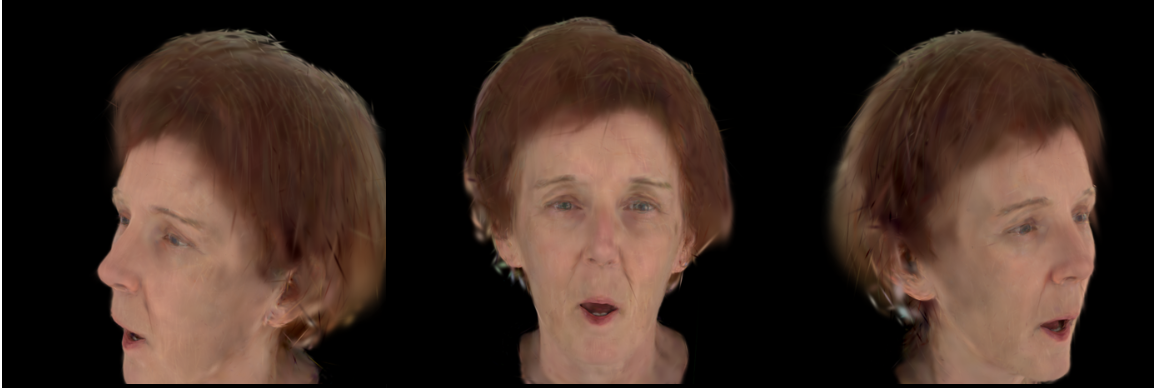}} \\
            \begin{tabular}{l}
                 \textbf{Ground} \\ \textbf{Truth}
            \end{tabular}  & \raisebox{-.5\height}{\includegraphics[width=0.6\linewidth]{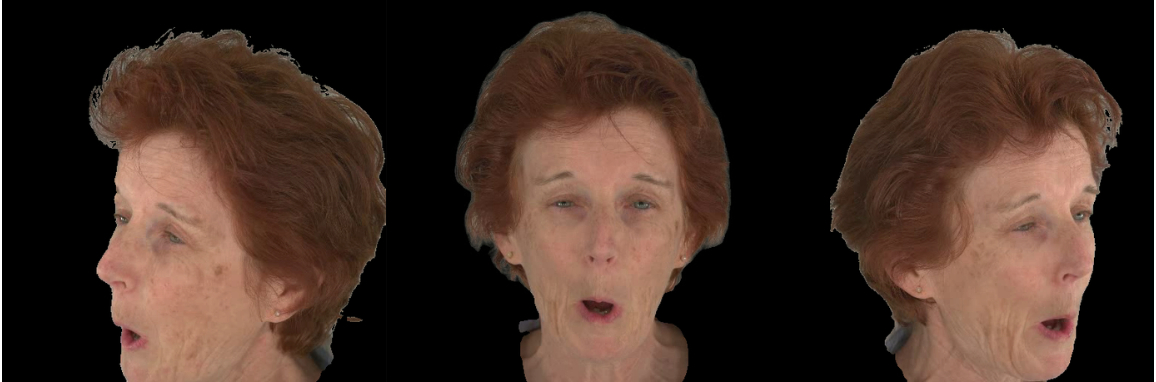}} \\
        \end{tabularx}
    \caption{Qualitative comparisons of our method with existing state-of-the-art in the \textbf{Single Image Setting}, using the top image only for the fitting process.}
    \label{fig:single_qual}
\end{figure}

The results of the single image setting are shown in \cref{tab:main_quant}. With such limited data, other Gaussian Avatar methods overfit and perform poorly. Even on the same camera view as the input image, Gaussian Avatar methods struggle with artefacts; see \cref{fig:single_qual}. Our method also outperforms ROME \cite{Khakhulin2022ROME}, which is designed to work with a single image. 

\subsection{Ablations}

We perform an ablation study to demonstrate our model's effectiveness. The results are shown in \cref{tab:ablations}. We use three subjects in the monocular setting. More details, as well as additional qualitative results, are in the supplementary. 

%\noindent\textbf
%\paragraph

\emph{No Prior:} To validate the use of the prior, we fit person-specific models using our MLP without any prior. We also ablate the use of the prior regularization loss term. It can be seen that the absence of the prior dramatically reduces the quality according to all metrics, while not regularizing towards the prior leads to slightly better ID reconstruction but worse quality according to all other metrics.

\emph{Number of Subjects:} We compare the model quality using priors trained on differing numbers of subjects. The more subjects we have, the better the quality. Interestingly, using one subject in the prior performs worse than not using a prior. This contrasts with the findings of \citet{buehler2024cafca}. 

\emph{Number of Gaussians:} We consider the effect of initializing with fewer Gaussians. We use texture maps ranging from $64\times64$ up to the full $512\times512$. We see that the highest resolution model performs best. Although the gain is small, it is notable visually (see supplementary). 

\emph{Fitting Stages:} We show the importance of each stage of the fitting process. Without stage 1 (optimizing for $\mathbf{z}$), our model performs worse on all metrics; this is also true for stage 2, although less pronounced. Without stage 3 our model performs similarly, or slightly better, visually, but suffers from a significant drop in ID reconstruction. The stages are seen visually in \cref{fig:stages_qual}.

\subsection{Runtime}

After fitting a user's Avatar using the prior, we can generate the mesh attached Gaussian Avatar parameters $\mathcal{A}$. Combined with the triangle face bindings and barycentric coordinates, this fully specifies an Avatar. No neural networks, including $\mathcal{D}$, are required for inference. A user's Avatar can be stored as an approximately 15MB file. Without any runtime optimizations, the complete inference pass, from Morphable Model parameters to the final rendered image runs at 70fps on an NVIDIA 4090 RTX GPU. The posing of the Gaussians can run at 67fps on a 3rd Gen Intel(R) Core(TM) i9-13900K CPU, suggesting improvements in Gaussian Splatting may allow real-time CPU inference.

\begin{table}[t]
\centering
\resizebox{\columnwidth}{!}{%
\begin{tabular}{l|cccccc}
\toprule
Method & PSNR $\uparrow$ & SSIM $\uparrow$ & LPIPS $\downarrow$ & FID $\downarrow$ & ID-SIM $\uparrow$ \\
\midrule
w/o prior & 19.42 & 0.670 & 0.391 & 212 & 0.478 \\
w/o prior regularization & 20.31 & 0.701 & 0.344 & 122 & \cellcolor{red!25}0.620 \\
\midrule
w/o stage 1 & 19.56 & 0.678 & 0.364 & 127 & 0.588 \\
w/o stage 2 & 20.33 & 0.704 & 0.347 & 118 & 0.585 \\
w/o stage 3 & \cellcolor{yellow!25}20.47 & \cellcolor{yellow!25}0.711 & \cellcolor{yellow!25}0.343 & \cellcolor{yellow!25}113 & 0.441 \\
\midrule
1 Prior Subject & 15.86 & 0.550 & 0.459 & 274 & 0.365 \\
10 Prior Subjects & 19.98 & 0.678 &  0.367 & 146 & 0.538 \\
100 Prior Subjects & 20.39 & 0.703 & 0.347 & 129 & 0.577 \\
\midrule
64$\times$64 Gaussians & 20.41 & 0.709 & 0.363 & 155 & 0.493 \\
128$\times$128 Gaussians & 20.45 & 0.704 & 0.353 & 127 & 0.567 \\
256$\times$256 Gaussians & 20.43 & 0.702 & 0.350 & 117 & 0.575 \\
\midrule
Full (1k Subjects, 512$\times$512) & \cellcolor{red!25}20.67 & \cellcolor{red!25}0.716 & \cellcolor{red!25}0.340 & \cellcolor{red!25}108 & \cellcolor{yellow!25}0.589 \\
\bottomrule
\end{tabular}
}
\caption{
\textbf{Ablations: } We ablate several components of the model. We evaluate the absence of the prior, the effect of fewer subjects and fewer Gaussians. We also ablate each stage of the fitting process. We highlight the \colorbox{red!25}{Best} and \colorbox{yellow!25}{Second Best} for each metric.
}
\label{tab:ablations}
\end{table}

\section{Limitations and Future Work}

While our model is able to achieve high-quality $360^\circ$ rendering, it has some limitations. For some regions, such as the back of the head, the model produces synthetic-looking results. We would like to address this issue by looking into 2D image-based priors \cite{Liu2023Zero1to3ZO, liu2023one2345++} based on diffusion models \cite{rombach2021highresolution}. To reduce artefacts introduced by overfitting to the monocular view, we used only flat RGB colour and did not model lighting, reducing our model's realism. In future, we may include a lighting model in our prior, enabled by a diverse set of lighting conditions in our synthetic data. As can be seen in our supplementary, our prior serves as a generative model with good interpretability. Given sufficient resources and a good camera/morphable model registration pipeline, we would like to use the findings of this work to train a similar generative prior using real data.
\section{Conclusion}

We have presented \textbf{GASP}, a novel method enabling 360$^\circ$, high-quality Avatar synthesis from limited data. Our model builds a prior over Gaussian Avatar parameters to ``fill in" missing regions. To bypass issues associated with collecting a large-scale real dataset, such as the need for full coverage and perfect annotation, we use synthetic data. Learned semantic Gaussian features and a three-stage fitting process enable us to cross the domain gap, while fitting to real data, to create realistic avatars. Our model outperforms the state-of-the-art in novel view and expression synthesis with Avatars trained from a single camera (e.g., a webcam or phone camera) using a short video or a single image while retaining the ability to animate and render in real-time.

\clearpage

{
    \small
    \bibliographystyle{ieeenat_fullname}
    \bibliography{main}

\begin{thebibliography}{47}
\providecommand{\natexlab}[1]{#1}
\providecommand{\url}[1]{\texttt{#1}}
\expandafter\ifx\csname urlstyle\endcsname\relax
  \providecommand{\doi}[1]{doi: #1}\else
  \providecommand{\doi}{doi: \begingroup \urlstyle{rm}\Url}\fi

\bibitem[Blanz and Vetter(2023)]{blanzvetter3DMM}
Volker Blanz and Thomas Vetter.
\newblock \emph{A Morphable Model For The Synthesis Of 3D Faces}.
\newblock Association for Computing Machinery, New York, NY, USA, 1 edition, 2023.

\bibitem[Buehler et~al.(2024)Buehler, Li, Wood, Helminger, Chen, Shah, Wang, Garbin, Orts-Escolano, Hilliges, Lagun, Riviere, Gotardo, Beeler, Meka, and Sarkar]{buehler2024cafca}
Marcel~C. Buehler, Gengyan Li, Erroll Wood, Leonhard Helminger, Xu Chen, Tanmay Shah, Daoye Wang, Stephan Garbin, Sergio Orts-Escolano, Otmar Hilliges, Dmitry Lagun, Jérémy Riviere, Paulo Gotardo, Thabo Beeler, Abhimitra Meka, and Kripasindhu Sarkar.
\newblock Cafca: High-quality novel view synthesis of expressive faces from casual few-shot captures.
\newblock In \emph{ACM SIGGRAPH Asia 2024 Conference Paper}. 2024.

\bibitem[B{\"u}hler et~al.(2023)B{\"u}hler, Sarkar, Shah, Li, Wang, Helminger, Orts-Escolano, Lagun, Hilliges, Beeler, et~al.]{buhler2023preface}
Marcel~C B{\"u}hler, Kripasindhu Sarkar, Tanmay Shah, Gengyan Li, Daoye Wang, Leonhard Helminger, Sergio Orts-Escolano, Dmitry Lagun, Otmar Hilliges, Thabo Beeler, et~al.
\newblock Preface: A data-driven volumetric prior for few-shot ultra high-resolution face synthesis.
\newblock In \emph{Proceedings of the IEEE/CVF International Conference on Computer Vision}, pages 3402--3413, 2023.

\bibitem[Chen et~al.(2023)Chen, Wang, Li, Xiao, Zhang, Yao, and Liu]{chen2023monogaussianavatar}
Yufan Chen, Lizhen Wang, Qijing Li, Hongjiang Xiao, Shengping Zhang, Hongxun Yao, and Yebin Liu.
\newblock Monogaussianavatar: Monocular gaussian point-based head avatar.
\newblock \emph{arXiv}, 2023.

\bibitem[Ding et~al.(2023)Ding, Xia, Jebe, Tu, and Zhang]{ding2023diffusionrig}
Cecilia Ding, Zheng ans~Zhang, Zhihao Xia, Lars Jebe, Zhuowen Tu, and Xiuming Zhang.
\newblock Diffusionrig: Learning personalized priors for facial appearance editing.
\newblock In \emph{Proceedings of the IEEE/CVF Conference on Computer Vision and Pattern Recognition}, 2023.

\bibitem[Gafni et~al.(2021)Gafni, Thies, Zollh{\"o}fer, and Nie{\ss}ner]{Gafni_2021_CVPR}
Guy Gafni, Justus Thies, Michael Zollh{\"o}fer, and Matthias Nie{\ss}ner.
\newblock Dynamic neural radiance fields for monocular 4d facial avatar reconstruction.
\newblock In \emph{Proceedings of the IEEE/CVF Conference on Computer Vision and Pattern Recognition (CVPR)}, pages 8649--8658, 2021.

\bibitem[Giebenhain et~al.(2023)Giebenhain, Kirschstein, Georgopoulos, R{\"{u}}nz, Agapito, and Nie{\ss}ner]{giebenhain2023nphm}
Simon Giebenhain, Tobias Kirschstein, Markos Georgopoulos, Martin R{\"{u}}nz, Lourdes Agapito, and Matthias Nie{\ss}ner.
\newblock Learning neural parametric head models.
\newblock In \emph{Proc. IEEE Conf. on Computer Vision and Pattern Recognition (CVPR)}, 2023.

\bibitem[Giebenhain et~al.(2024)Giebenhain, Kirschstein, R{\"{u}}nz, Agapito, and Nie{\ss}ner]{giebenhain2024npga}
Simon Giebenhain, Tobias Kirschstein, Martin R{\"{u}}nz, Lourdes Agapito, and Matthias Nie{\ss}ner.
\newblock Npga: Neural parametric gaussian avatars.
\newblock In \emph{SIGGRAPH Asia 2024 Conference Papers (SA Conference Papers '24), December 3-6, Tokyo, Japan}, 2024.

\bibitem[Guo et~al.(2021)Guo, Chen, Liang, Liu, Bao, and Zhang]{guo2021adnerf}
Yudong Guo, Keyu Chen, Sen Liang, Yongjin Liu, Hujun Bao, and Juyong Zhang.
\newblock Ad-nerf: Audio driven neural radiance fields for talking head synthesis.
\newblock In \emph{IEEE/CVF International Conference on Computer Vision (ICCV)}, 2021.

\bibitem[Hearst et~al.(1998)Hearst, Dumais, Osuna, Platt, and Scholkopf]{hearst1998support}
Marti~A. Hearst, Susan~T Dumais, Edgar Osuna, John Platt, and Bernhard Scholkopf.
\newblock Support vector machines.
\newblock \emph{IEEE Intelligent Systems and their applications}, 13\penalty0 (4):\penalty0 18--28, 1998.

\bibitem[Heusel et~al.(2017)Heusel, Ramsauer, Unterthiner, Nessler, and Hochreiter]{FID17}
Martin Heusel, Hubert Ramsauer, Thomas Unterthiner, Bernhard Nessler, and Sepp Hochreiter.
\newblock Gans trained by a two time-scale update rule converge to a local nash equilibrium.
\newblock In \emph{Proceedings of the 31st International Conference on Neural Information Processing Systems}, page 6629–6640, Red Hook, NY, USA, 2017. Curran Associates Inc.

\bibitem[Hewitt et~al.(2024)Hewitt, Saleh, Aliakbarian, Petikam, Rezaeifar, Florentin, Hosenie, Cashman, Valentin, Cosker, and Baltru\v{s}aitis]{hewitt2024look}
Charlie Hewitt, Fatemeh Saleh, Sadegh Aliakbarian, Lohit Petikam, Shideh Rezaeifar, Louis Florentin, Zafiirah Hosenie, Thomas~J Cashman, Julien Valentin, Darren Cosker, and Tadas Baltru\v{s}aitis.
\newblock Look ma, no markers: holistic performance capture without the hassle.
\newblock \emph{ACM Transactions on Graphics (TOG)}, 36\penalty0 (6), 2024.

\bibitem[Ho et~al.(2020)Ho, Jain, and Abbeel]{ho2020denoising}
Jonathan Ho, Ajay Jain, and Pieter Abbeel.
\newblock Denoising diffusion probabilistic models.
\newblock \emph{arXiv preprint arxiv:2006.11239}, 2020.

\bibitem[Karras et~al.(2020)Karras, Laine, Aittala, Hellsten, Lehtinen, and Aila]{Karras2019stylegan2}
Tero Karras, Samuli Laine, Miika Aittala, Janne Hellsten, Jaakko Lehtinen, and Timo Aila.
\newblock Analyzing and improving the image quality of {StyleGAN}.
\newblock In \emph{Proc. CVPR}, 2020.

\bibitem[Kerbl et~al.(2023)Kerbl, Kopanas, Leimk{\"u}hler, and Drettakis]{kerbl3Dgaussians}
Bernhard Kerbl, Georgios Kopanas, Thomas Leimk{\"u}hler, and George Drettakis.
\newblock 3d gaussian splatting for real-time radiance field rendering.
\newblock \emph{ACM Transactions on Graphics}, 42\penalty0 (4), 2023.

\bibitem[Khakhulin et~al.(2022)Khakhulin, Sklyarova, Lempitsky, and Zakharov]{Khakhulin2022ROME}
Taras Khakhulin, Vanessa Sklyarova, Victor Lempitsky, and Egor Zakharov.
\newblock Realistic one-shot mesh-based head avatars.
\newblock In \emph{European Conference of Computer vision (ECCV)}, 2022.

\bibitem[Kim et~al.(2018)Kim, Garrido, Tewari, Xu, Thies, Nie{\ss}ner, P{\'e}rez, Richardt, Zoll{\"o}fer, and Theobalt]{kim2018deep}
Hyeongwoo Kim, Pablo Garrido, Ayush Tewari, Weipeng Xu, Justus Thies, Matthias Nie{\ss}ner, Patrick P{\'e}rez, Christian Richardt, Michael Zoll{\"o}fer, and Christian Theobalt.
\newblock Deep video portraits.
\newblock \emph{ACM Transactions on Graphics (TOG)}, 37\penalty0 (4):\penalty0 163, 2018.

\bibitem[Kingma and Ba(2014)]{Kingma2014AdamAM}
Diederik~P. Kingma and Jimmy Ba.
\newblock Adam: A method for stochastic optimization.
\newblock \emph{CoRR}, abs/1412.6980, 2014.

\bibitem[Kirschstein et~al.(2023)Kirschstein, Qian, Giebenhain, Walter, and Nie\ss{}ner]{kirschstein2023nersemble}
Tobias Kirschstein, Shenhan Qian, Simon Giebenhain, Tim Walter, and Matthias Nie\ss{}ner.
\newblock Nersemble: Multi-view radiance field reconstruction of human heads.
\newblock \emph{ACM Trans. Graph.}, 42\penalty0 (4), 2023.

\bibitem[Li et~al.(2017)Li, Bolkart, Black, Li, and Romero]{FLAME:SiggraphAsia2017}
Tianye Li, Timo Bolkart, Michael.~J. Black, Hao Li, and Javier Romero.
\newblock Learning a model of facial shape and expression from {4D} scans.
\newblock \emph{ACM Transactions on Graphics, (Proc. SIGGRAPH Asia)}, 36\penalty0 (6):\penalty0 194:1--194:17, 2017.

\bibitem[Lin et~al.(2021)Lin, Yang, Saleemi, and Sengupta]{rvm}
Shanchuan Lin, Linjie Yang, Imran Saleemi, and Soumyadip Sengupta.
\newblock Robust high-resolution video matting with temporal guidance, 2021.

\bibitem[Liu et~al.(2023{\natexlab{a}})Liu, Shi, Chen, Zhang, Xu, Wei, Chen, Zeng, Gu, and Su]{liu2023one2345++}
Minghua Liu, Ruoxi Shi, Linghao Chen, Zhuoyang Zhang, Chao Xu, Xinyue Wei, Hansheng Chen, Chong Zeng, Jiayuan Gu, and Hao Su.
\newblock One-2-3-45++: Fast single image to 3d objects with consistent multi-view generation and 3d diffusion.
\newblock \emph{arXiv preprint arXiv:2311.07885}, 2023{\natexlab{a}}.

\bibitem[Liu et~al.(2023{\natexlab{b}})Liu, Wu, Hoorick, Tokmakov, Zakharov, and Vondrick]{Liu2023Zero1to3ZO}
Ruoshi Liu, Rundi Wu, Basile~Van Hoorick, Pavel Tokmakov, Sergey Zakharov, and Carl Vondrick.
\newblock Zero-1-to-3: Zero-shot one image to 3d object.
\newblock \emph{2023 IEEE/CVF International Conference on Computer Vision (ICCV)}, pages 9264--9275, 2023{\natexlab{b}}.

\bibitem[Ma et~al.(2021)Ma, Simon, Saragih, Wang, Li, Torre, and Sheikh]{CodecAvatars}
S. Ma, T. Simon, J. Saragih, D. Wang, Y. Li, F.~La Torre, and Y. Sheikh.
\newblock Pixel codec avatars.
\newblock In \emph{2021 IEEE/CVF Conference on Computer Vision and Pattern Recognition (CVPR)}, pages 64--73, Los Alamitos, CA, USA, 2021. IEEE Computer Society.

\bibitem[Mildenhall et~al.(2020{\natexlab{a}})Mildenhall, Srinivasan, Tancik, Barron, Ramamoorthi, and Ng]{Mildenhall20eccv_nerf}
Ben Mildenhall, Pratul~P. Srinivasan, Matthew Tancik, Jonathan~T. Barron, Ravi Ramamoorthi, and Ren Ng.
\newblock {NeRF}: Representing scenes as neural radiance fields for view synthesis.
\newblock In \emph{The European Conference on Computer Vision (ECCV)}, 2020{\natexlab{a}}.

\bibitem[Mildenhall et~al.(2020{\natexlab{b}})Mildenhall, Srinivasan, Tancik, Barron, Ramamoorthi, and Ng]{mildenhall2020nerf}
Ben Mildenhall, Pratul~P. Srinivasan, Matthew Tancik, Jonathan~T. Barron, Ravi Ramamoorthi, and Ren Ng.
\newblock Nerf: Representing scenes as neural radiance fields for view synthesis, 2020{\natexlab{b}}.

\bibitem[Park et~al.(2019)Park, Florence, Straub, Newcombe, and Lovegrove]{Park_2019_CVPR}
Jeong~Joon Park, Peter Florence, Julian Straub, Richard Newcombe, and Steven Lovegrove.
\newblock Deepsdf: Learning continuous signed distance functions for shape representation.
\newblock In \emph{The IEEE Conference on Computer Vision and Pattern Recognition (CVPR)}, 2019.

\bibitem[Qian et~al.(2023)Qian, Kirschstein, Schoneveld, Davoli, Giebenhain, and Nie\ss{}ner]{qian2023gaussianavatars}
Shenhan Qian, Tobias Kirschstein, Liam Schoneveld, Davide Davoli, Simon Giebenhain, and Matthias Nie\ss{}ner.
\newblock Gaussianavatars: Photorealistic head avatars with rigged 3d gaussians.
\newblock \emph{arXiv preprint arXiv:2312.02069}, 2023.

\bibitem[Rao et~al.(2022)Rao, R, Fox, Weyrich, Bickel, Pfister, Matusik, Tewari, Theobalt, and Elgharib]{Rao_2022_BMVC}
Pramod Rao, Mallikarjun~B R, Gereon Fox, Tim Weyrich, Bernd Bickel, Hanspeter Pfister, Wojciech Matusik, Ayush Tewari, Christian Theobalt, and Mohamed Elgharib.
\newblock Vorf: Volumetric relightable faces.
\newblock In \emph{33rd British Machine Vision Conference 2022, {BMVC} 2022, London, UK, November 21-24, 2022}. {BMVA} Press, 2022.

\bibitem[Rombach et~al.(2021)Rombach, Blattmann, Lorenz, Esser, and Ommer]{rombach2021highresolution}
Robin Rombach, Andreas Blattmann, Dominik Lorenz, Patrick Esser, and Björn Ommer.
\newblock High-resolution image synthesis with latent diffusion models, 2021.

\bibitem[Saunders and Namboodiri(2024)]{Saunders2024D4E}
Jack Saunders and Vinay Namboodiri.
\newblock Dubbing for everyone: Data-efficient visual dubbing using neural rendering priors.
\newblock \emph{arxiv}, 2024.

\bibitem[Saunders and Namboodiri(2023)]{Saunders2023READ}
Jack Saunders and Vinay~P. Namboodiri.
\newblock Read avatars: Realistic emotion-controllable audio driven avatars.
\newblock In \emph{arxiv}, 2023.

\bibitem[Shao et~al.(2024)Shao, Wang, Li, Wang, Lin, Zhang, Fan, and Wang]{shao2024splattingavatar}
Zhijing Shao, Zhaolong Wang, Zhuang Li, Duotun Wang, Xiangru Lin, Yu Zhang, Mingming Fan, and Zeyu Wang.
\newblock {SplattingAvatar: Realistic Real-Time Human Avatars with Mesh-Embedded Gaussian Splatting}.
\newblock In \emph{Proceedings of the IEEE/CVF Conference on Computer Vision and Pattern Recognition (CVPR)}, 2024.

\bibitem[Tewari et~al.(2020)Tewari, Elgharib, Bharaj, Bernard, Seidel, Pérez, Zollhöfer, and Theobalt]{StyleRig}
Ayush Tewari, Mohamed Elgharib, Gaurav Bharaj, Florian Bernard, Hans-Peter Seidel, Patrick Pérez, Michael Zollhöfer, and Christian Theobalt.
\newblock Stylerig: Rigging stylegan for 3d control over portrait images.
\newblock In \emph{2020 IEEE/CVF Conference on Computer Vision and Pattern Recognition (CVPR)}, pages 6141--6150, 2020.

\bibitem[Thies et~al.(2018)Thies, Zollh\"{o}fer, Stamminger, Theobalt, and Nie\ss{}ner]{Face2Face}
Justus Thies, Michael Zollh\"{o}fer, Marc Stamminger, Christian Theobalt, and Matthias Nie\ss{}ner.
\newblock Face2face: real-time face capture and reenactment of rgb videos.
\newblock \emph{Commun. ACM}, 62\penalty0 (1):\penalty0 96–104, 2018.

\bibitem[Thies et~al.(2019)Thies, Zollh\"{o}fer, and Nie\ss{}ner]{DNR}
Justus Thies, Michael Zollh\"{o}fer, and Matthias Nie\ss{}ner.
\newblock Deferred neural rendering: image synthesis using neural textures.
\newblock \emph{ACM Trans. Graph.}, 38\penalty0 (4), 2019.

\bibitem[Thies et~al.(2020)Thies, Elgharib, Tewari, Theobalt, and Nie{\ss}ner]{thies2020nvp}
Justus Thies, Mohamed Elgharib, Ayush Tewari, Christian Theobalt, and Matthias Nie{\ss}ner.
\newblock Neural voice puppetry: Audio-driven facial reenactment.
\newblock \emph{ECCV 2020}, 2020.

\bibitem[Wood et~al.(2021)Wood, Baltru{\v{s}}aitis, Hewitt, Dziadzio, Cashman, and Shotton]{wood2021fake}
Erroll Wood, Tadas Baltru{\v{s}}aitis, Charlie Hewitt, Sebastian Dziadzio, Thomas~J Cashman, and Jamie Shotton.
\newblock Fake it till you make it: face analysis in the wild using synthetic data alone.
\newblock In \emph{Proceedings of the IEEE/CVF international conference on computer vision}, pages 3681--3691, 2021.

\bibitem[Wuu et~al.(2022)Wuu, Zheng, Ardisson, Bali, Belko, Brockmeyer, Evans, Godisart, Ha, Huang, Hypes, Koska, Krenn, Lombardi, Luo, McPhail, Millerschoen, Perdoch, Pitts, Richard, Saragih, Saragih, Shiratori, Simon, Stewart, Trimble, Weng, Whitewolf, Wu, Yu, and Sheikh]{wuu2022multiface}
Cheng-hsin Wuu, Ningyuan Zheng, Scott Ardisson, Rohan Bali, Danielle Belko, Eric Brockmeyer, Lucas Evans, Timothy Godisart, Hyowon Ha, Xuhua Huang, Alexander Hypes, Taylor Koska, Steven Krenn, Stephen Lombardi, Xiaomin Luo, Kevyn McPhail, Laura Millerschoen, Michal Perdoch, Mark Pitts, Alexander Richard, Jason Saragih, Junko Saragih, Takaaki Shiratori, Tomas Simon, Matt Stewart, Autumn Trimble, Xinshuo Weng, David Whitewolf, Chenglei Wu, Shoou-I Yu, and Yaser Sheikh.
\newblock Multiface: A dataset for neural face rendering.
\newblock In \emph{arXiv}, 2022.

\bibitem[Xiang et~al.(2024)Xiang, Gao, Guo, and Zhang]{xiang2024flashavatar}
Jun Xiang, Xuan Gao, Yudong Guo, and Juyong Zhang.
\newblock Flashavatar: High-fidelity head avatar with efficient gaussian embedding.
\newblock In \emph{The IEEE Conference on Computer Vision and Pattern Recognition (CVPR)}, 2024.

\bibitem[Xu et~al.(2024{\natexlab{a}})Xu, Chen, Li, Zhang, Wang, Zheng, and Liu]{xu2023gaussianheadavatar}
Yuelang Xu, Benwang Chen, Zhe Li, Hongwen Zhang, Lizhen Wang, Zerong Zheng, and Yebin Liu.
\newblock Gaussian head avatar: Ultra high-fidelity head avatar via dynamic gaussians.
\newblock In \emph{Proceedings of the IEEE/CVF Conference on Computer Vision and Pattern Recognition (CVPR)}, 2024{\natexlab{a}}.

\bibitem[Xu et~al.(2024{\natexlab{b}})Xu, Wang, Zheng, Su, and Liu]{xu2023gphm}
Yuelang Xu, Lizhen Wang, Zerong Zheng, Zhaoqi Su, and Yebin Liu.
\newblock 3d gaussian parametric head model.
\newblock In \emph{Proceedings of the European Conference on Computer Vision (ECCV)}, 2024{\natexlab{b}}.

\bibitem[Ye et~al.(2023)Ye, Jiang, Ren, Liu, He, and Zhao]{ye2023geneface}
Zhenhui Ye, Ziyue Jiang, Yi Ren, Jinglin Liu, Jinzheng He, and Zhou Zhao.
\newblock Geneface: Generalized and high-fidelity audio-driven 3d talking face synthesis.
\newblock \emph{arXiv preprint arXiv:2301.13430}, 2023.

\bibitem[Yu et~al.(2021)Yu, Gao, Wang, Yu, Shen, and Sang]{BiSeNet}
Changqian Yu, Changxin Gao, Jingbo Wang, Gang Yu, Chunhua Shen, and Nong Sang.
\newblock Bisenet v2: Bilateral network with guided aggregation for real-time semantic segmentation.
\newblock \emph{Int. J. Comput. Vision}, 129\penalty0 (11):\penalty0 3051–3068, 2021.

\bibitem[Zhang et~al.(2018)Zhang, Isola, Efros, Shechtman, and Wang]{zhang2018perceptual}
Richard Zhang, Phillip Isola, Alexei~A Efros, Eli Shechtman, and Oliver Wang.
\newblock The unreasonable effectiveness of deep features as a perceptual metric.
\newblock In \emph{CVPR}, 2018.

\bibitem[Zheng et~al.(2023)Zheng, Yifan, Wetzstein, Black, and Hilliges]{Zheng2023pointavatar}
Yufeng Zheng, Wang Yifan, Gordon Wetzstein, Michael~J. Black, and Otmar Hilliges.
\newblock Pointavatar: Deformable point-based head avatars from videos.
\newblock In \emph{Proceedings of the IEEE/CVF Conference on Computer Vision and Pattern Recognition (CVPR)}, 2023.

\bibitem[Zielonka et~al.(2022)Zielonka, Bolkart, and Thies]{Zielonka2022InstantVH}
Wojciech Zielonka, Timo Bolkart, and Justus Thies.
\newblock Instant volumetric head avatars.
\newblock \emph{2023 IEEE/CVF Conference on Computer Vision and Pattern Recognition (CVPR)}, pages 4574--4584, 2022.

\end{thebibliography}
}

\newpage
\appendix
\clearpage
\setcounter{page}{1}
\maketitlesupplementary

\section{Further Results}

\begin{figure*}
    \centering
    \includegraphics[width=\textwidth]{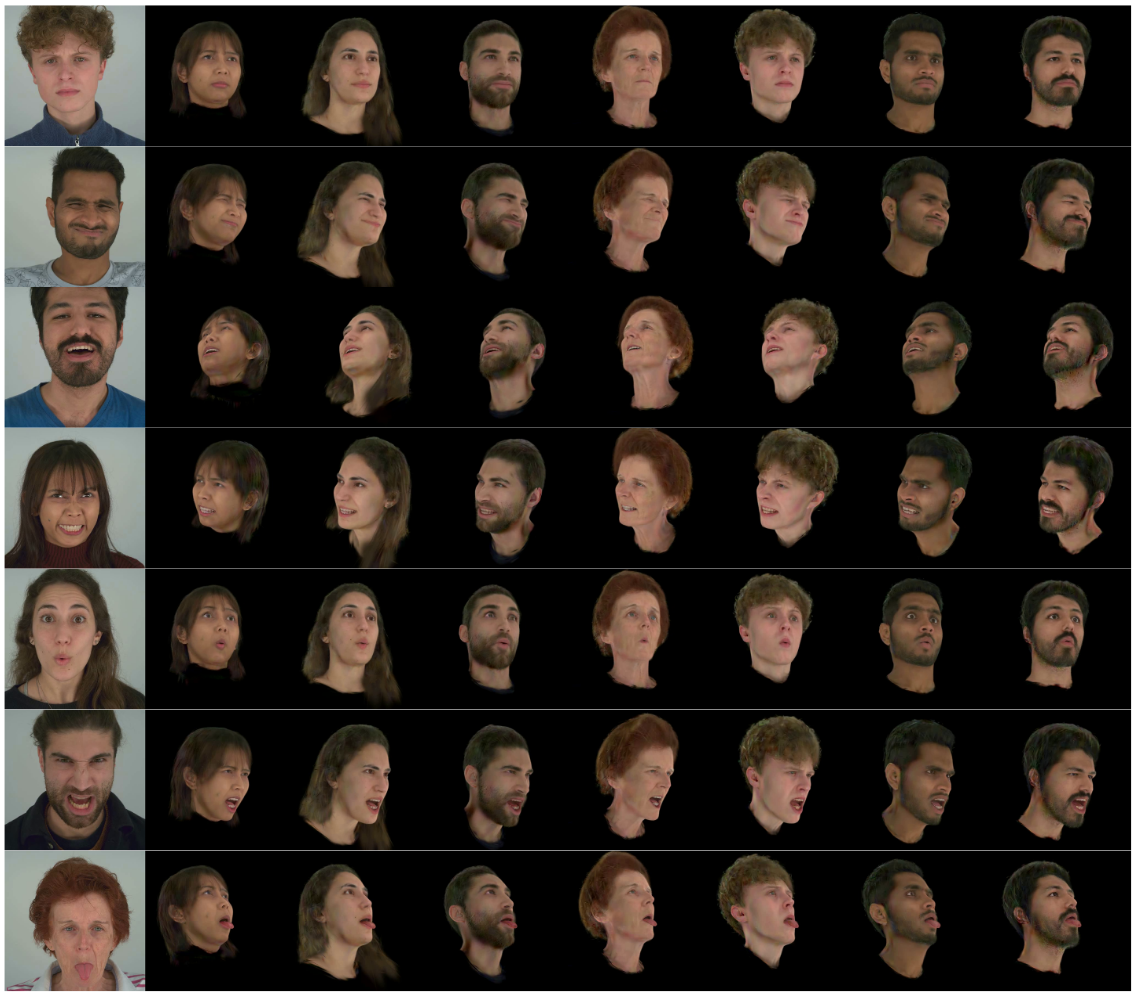}
    \caption{\textbf{Additional Cross Reenactment Results}. We show several more examples of cross-reenactment. We use the input image on the left to drive the avatars on the right. Each Avatar is trained in the \textbf{Monocular Setting}.}
    \label{fig:more_cross}
\end{figure*}

\begin{figure*}
    \centering
    \includegraphics[width=\textwidth]{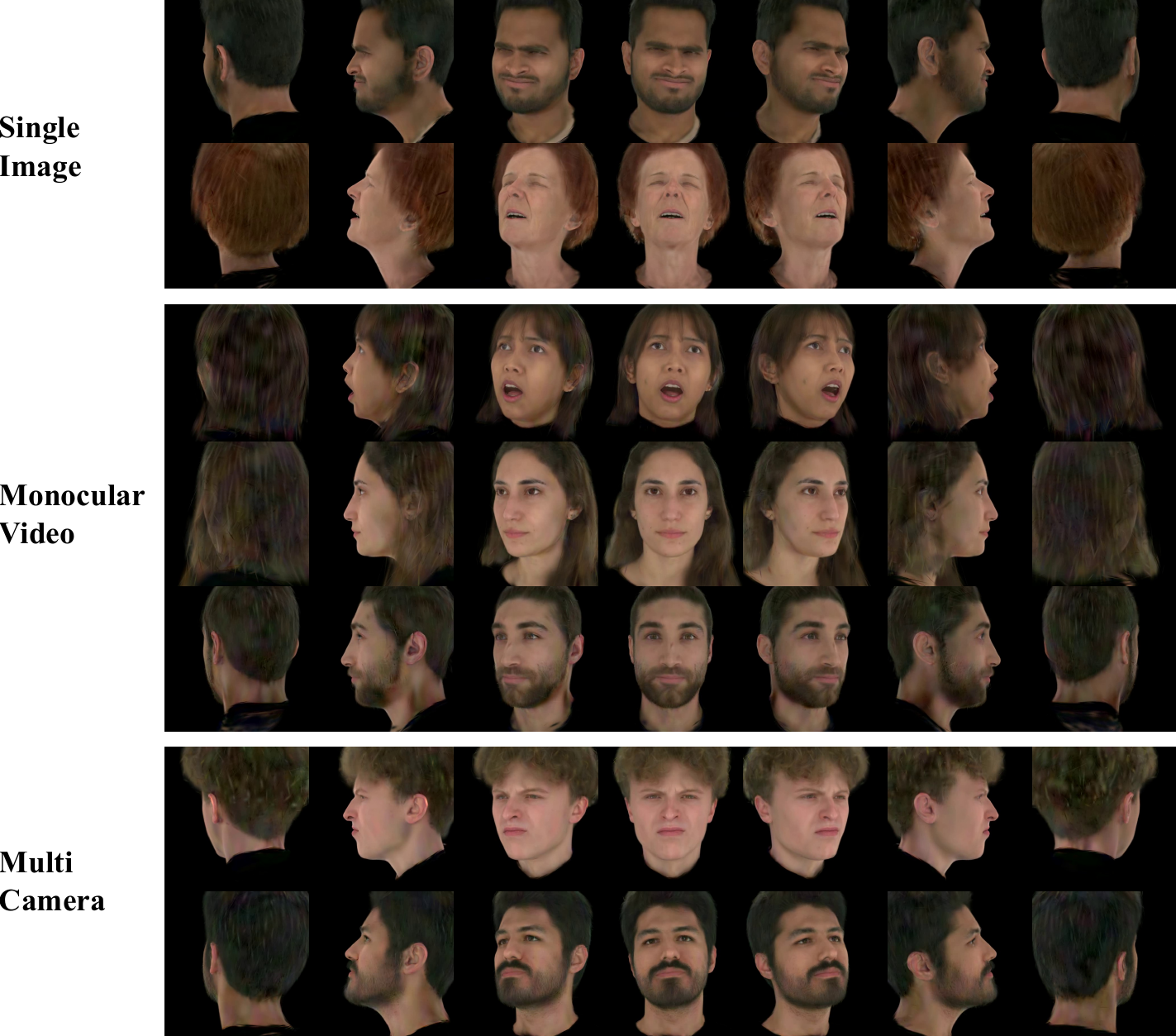}
    \caption{\textbf{Additional Self Reenactment Results}. We show several more examples of self-reenactment with 360$^\circ$ rendering. We show models fit to a single image (Top), a monocular video (Middle) and multiple views (Bottom). In each case, the back of the head is never included in the fitting data.}
    \label{fig:more_self}
\end{figure*}

We show further examples of self-reenactment, wherein we take an unseen video of the subject and use it to drive their avatar. We show full 360$^\circ$ renderings of the head. The results are shown in \cref{fig:more_self}. Despite having never seen the back of an actual person's head, our model produces plausible results. We also show cross-identity re-enactment, taking a video from one Avatar to animate several others. This is demonstrated in \cref{fig:more_cross}. Video versions of these results are also shown in our supplemental video. 

\section{Latent space controllability}
\begin{figure*}
    \centering
    \includegraphics[width=\textwidth]{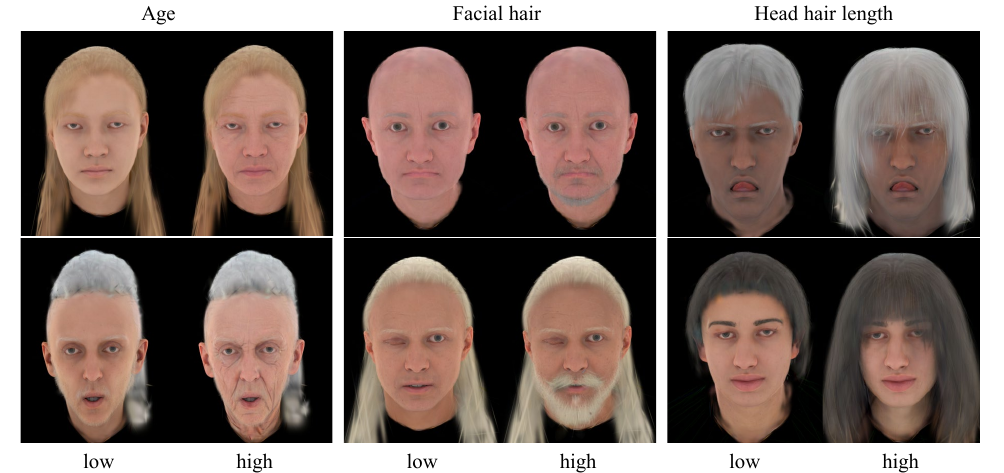}
    \caption{We demonstrate that the latent space learned by our prior model is controllable by finding directions in it that correspond to semantic features such as age, facial hair and hair length.}
    \label{fig:controllability}
\end{figure*}
To demonstrate that our prior model learns a controllable latent space, we propose a simple method for finding directions $\mathbf{d}_k$ in the latent space that are semantically meaningful. We then demonstrate that adding or subtracting those direction from a given identity's latent vector $\mathbf{z}_j$ leads to the desired changes in the person's appearance. The results of this process are shown in Figure \ref{fig:controllability}.

To learn $\mathbf{d}_k$ for a given semantic feature, we group our training data into samples that have this feature and samples that do not have it. As our training data is synthetic and extensively labeled, doing so is a matter of checking the metadata of the samples. We then take a pre-trained prior model and extract the $\mathbf{z}_j$ for each training sample. Finally, we train a Linear Support Vector Machine \cite{hearst1998support} that classifies the training data samples into ones that have the semantic feature and ones that do not have it, given the sample's $\mathbf{z}_j$. The direction $\mathbf{d}_k$ estimated by the Linear SVM is a vector orthogonal to the hyperplane that separate the two groups in the latent space of the prior model. Thus, adding $\mathbf{d}_k$ to a sample's latent vector $\mathbf{z}_j$ should move it closer to samples that have the feature, and subtracting it should have the opposite effect.

We evaluate this approach on three features:
\begin{enumerate}
\item Age - this corresponds to the age of the person whose facial texture was used in the training data sample. The SVM here was learned to classify age $\geq45$ into a separate group from age $<45$.
\item Facial hair - here, the SVM was learned to classify samples with facial hair separately from samples with no facial hair.
\item Head hair - here, the SVM separated samples with long hair from samples with short hair.
\end{enumerate}
The results of the evaluation are shown in the supplementary video as well as in Figure \ref{fig:controllability}, where each column demonstrates one of the features we control.

\section{MLP Architecture}

\begin{figure}
    \centering
    \includegraphics[width=0.5\textwidth]{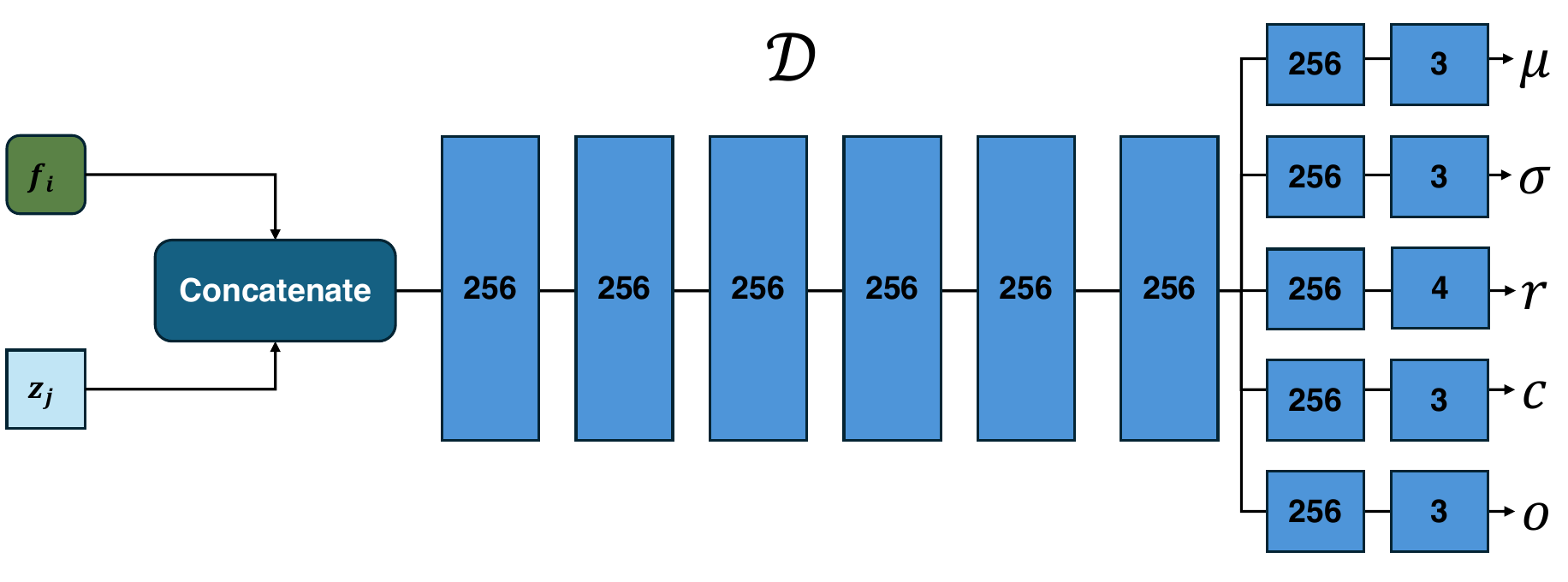}
    \caption{The architecture of our MLP decoder $\mathcal{D}$. $\mathbf{f}$ and $\mathcal{z}$ are concatenated and passed through 6 linear layers with output size 256. The network then splits into per-attribute branches. Each block represents a linear layer followed by ReLU and using weight normalization.}
    \label{fig:MLP_arch}
\end{figure}

Here, we give more detail about the architecture of our MLP Decoder $\mathcal{D}$. The network takes each 8-dimensional Gaussian feature $\mathbf{f_i}$ as input and concatenates them with the 512-dimensional vector $\mathbf{z_j}$ for the identity of the Avatar. This gives a 512-dimensional vector. These inputs are then passed through six linear layers with an output dimensionality 256. After this, the network separates into separate branches for position $\mu$, scale $\sigma$, rotation $r$, color $c$ and opacity $o$. Each branch has one linear layer with output dimension 256, followed by a final linear projection to the relevant shape for that attribute. Each linear layer, except the final projection, is followed by the ReLU activation function. Weight normalization is used on each layer. We visualise this architecture in \cref{fig:MLP_arch}

\section{Experimental Setup}

\label{sec:exp}

\begin{table}
    \centering
    \begin{tabular}{c|c?c|c}
    \toprule
        Subject &  Test Cameras & Subject & Test Cameras \\
    \midrule
        36 &\begin{tabular}{@{}c@{}c@{}c@{}}221501007 \\ 222200040 \\ 222200044 \\ 222200046 \end{tabular} & 37 &\begin{tabular}{@{}c@{}c@{}c@{}}221501007 \\ 222200040 \\ 222200044 \\ 222200045 \end{tabular} \\
        \hline
        57 &\begin{tabular}{@{}c@{}c@{}c@{}} 221501007 \\ 222200040 \\ 222200044 \\ 222200046 \end{tabular} & 74 &\begin{tabular}{@{}c@{}c@{}c@{}} 221501007 \\ 222200040 \\ 222200042 \\  222200044 \end{tabular} \\
        \hline
        100 &\begin{tabular}{@{}c@{}c@{}c@{}}221501007 \\ 222200039 \\ 222200042 \\ 222200045 \end{tabular} & 145 &\begin{tabular}{@{}c@{}c@{}c@{}} 221501007 \\ 222200042 \\ 222200044 \\ 222200045 \end{tabular} \\
        \hline
        165 &\begin{tabular}{@{}c@{}c@{}c@{}}221501007 \\ 222200042 \\ 222200044 \\ 222200045 \end{tabular} & 251 &\begin{tabular}{@{}c@{}c@{}c@{}}221501007 \\ 222200042 \\ 222200044 \\ 222200045\end{tabular} \\

    \bottomrule
    \end{tabular}
    \caption{Cameras selected as the most extreme view for each subject. The selection was performed empirically.}
    \label{tab:test_cameras}
\end{table}

Here, we discuss the exact setup of the experiments in the main paper. Recall we consider three experimental setups: Monocular, Single Frame and Multi Camera.

\paragraph{Monocular} For each training subject, we used the following sequences as training data:  EMO-1-shout+laugh, EMO-2-surprise+fear, EMO-3-angry+sad, EMO-4-disgust+happy, EXP-2-eyes, EXP-3-cheeks+nose, EXP-4-lips, EXP-5-mouth, EXP-6-tongue-1, EXP-7-tongue-2, EXP-8-jaw-1, EXP-9-jaw-2. For all subjects except 57, the camera 222200037 was selected as the most frontal, for subject 57 this was 222200038. These are the cameras we used in training. We subsample every other frame.

\paragraph{Single Image} For each training subject, we used the first frame of the sequence EMO-1-shout+laugh for training data. For all subjects except 57, the camera 222200037 was selected as the most frontal, for subject 57 this was 222200038. These are the cameras we used in training.

\paragraph{Multi Camera} For each training subject, we used the following sequences as training data:  EMO-1-shout+laugh, EMO-2-surprise+fear, EMO-3-angry+sad, EMO-4-disgust+happy, EXP-2-eyes, EXP-3-cheeks+nose, EXP-4-lips, EXP-5-mouth, EXP-6-tongue-1, EXP-7-tongue-2, EXP-8-jaw-1, EXP-9-jaw-2. We used all 16 Cameras. In order to reduce the size of these datasets, we subsampled every 10th frame, effectively taking each video at 7fps.

\paragraph{Testing} Testing on all subjects was performed using the FREE sequence, which has no overlap with any of our training sets. We used cameras as shown in \cref{tab:test_cameras}. For the main quantitivate results, we subsample every 5th frame to reduce computational overhead. For generating the qualitative videos we use every frame of the FREE sequence.

\section{Three Frame Model}

\begin{figure*}
    \centering
    \includegraphics[width=0.8\textwidth]{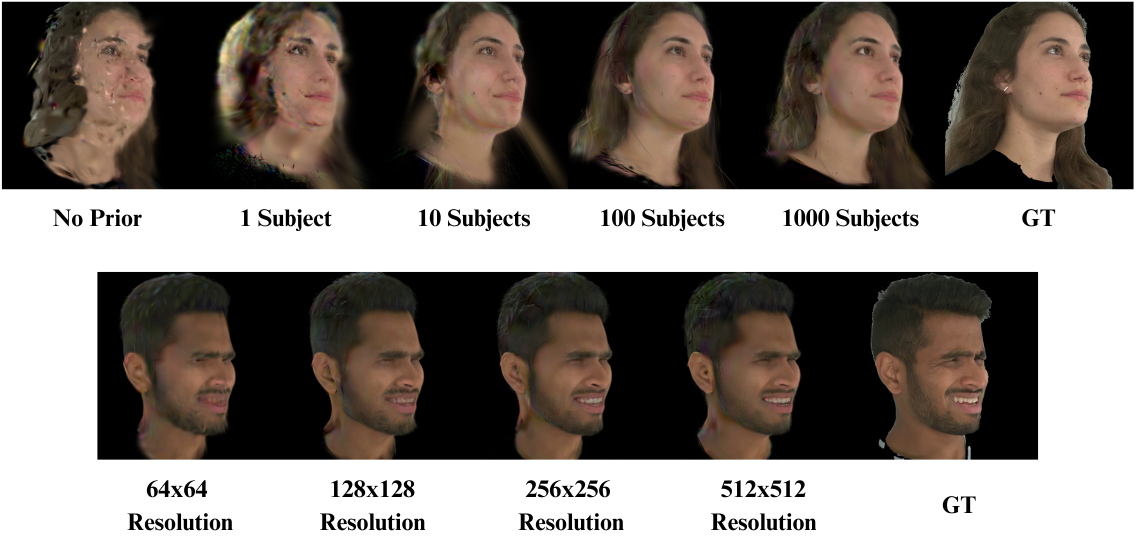}
    \caption{\textbf{Ablations: } We show the qualitative effect of using differing numbers of subjects to train the prior (top) and different numbers of Gaussians (bottom).}
    \label{fig:ablation_qual}
\end{figure*}

\begin{figure}
    \includegraphics[width=0.9\columnwidth]{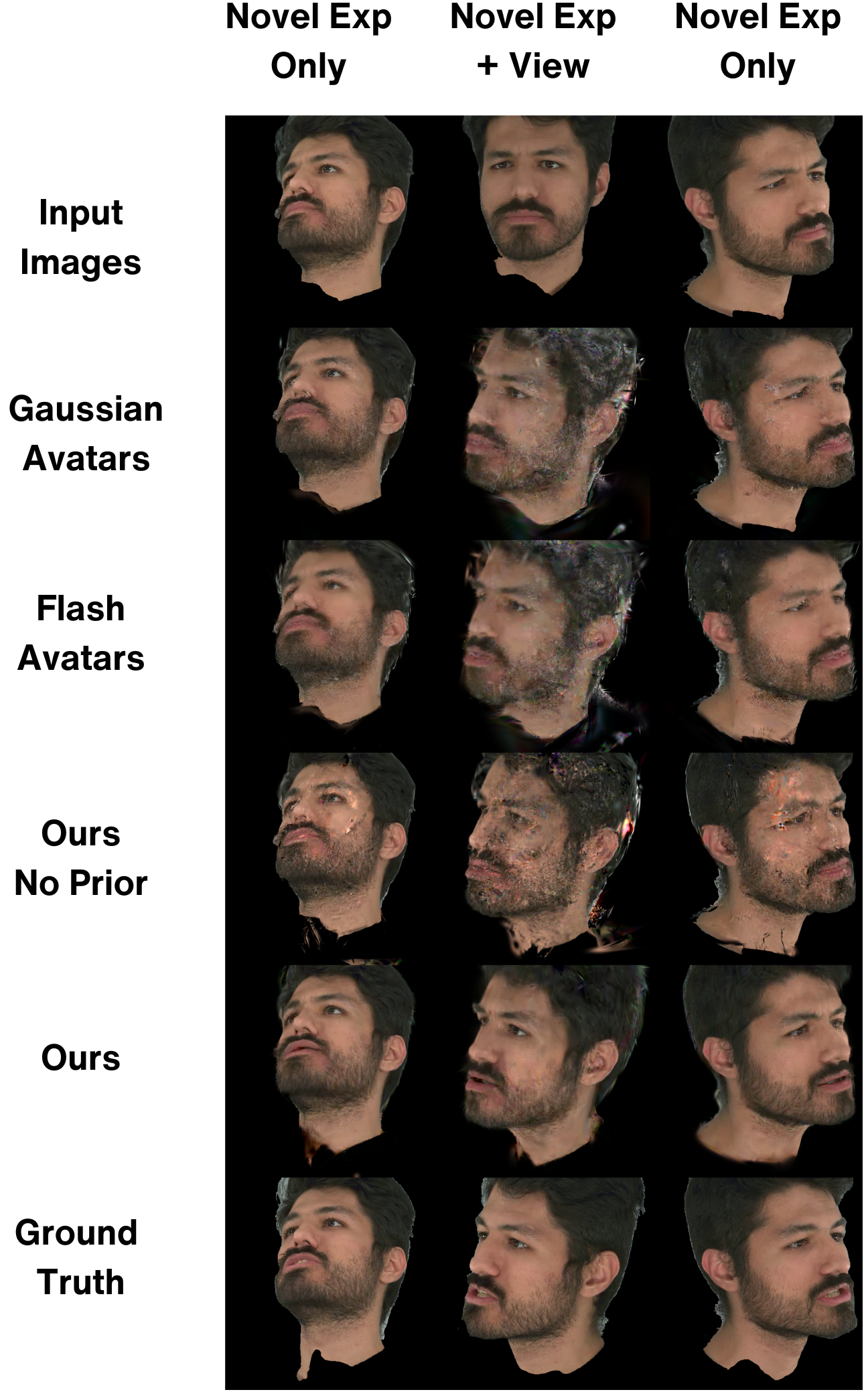}
    \caption{Qualitative comparisons of our method with existing state-of-the-art in the \textbf{Three Image Setting}, using the top 3 images as input. We show both novel expression and novel view synthesis in this setup.}
    \label{fig:three_qual}
\end{figure}

Some other few-shot Avatar models (e.g., \cite{buhler2023preface, buehler2024cafca}) address a related but different experimental setup using three frames, one frontal facing, one from the left and one from the right. While these models are not available for comparison, we replicate their setup here. For this, we select one image from the front, left and right of a model. We show some of the results in \cref{fig:three_qual}. It can be seen here that our model performs somewhat better on novel expressions from one of the training views (the left and right columns) and has significantly fewer artefacts on a novel view (middle column).

\section{User Study}

For our user study, we ask participants to rate the quality of each method. We show each method the FREE sequence played from the four extreme test cameras in \cref{tab:test_cameras}. Each participant is shown each combination of method and training setting (Monocular, Single Frame and Multi-Camera) for an individual subject, meaning a total of 13 images per user (Four methods times 3 settings plus the Single Frame setting for ROME \cite{Khakhulin2022ROME}). Images are shown in a grid of two-by-two using each of the four camera angles. We perform this for each of the eight test subjects we have run evaluation on, with users being assigned one of these subjects at random. Video order is also randomized to prevent bias. We conducted the user study using Amazon's Mechanical Turk. In total, 40 users completed the user study. The results are shown in \cref{tab:main_quant} and \cref{tab:multicam_quant}.

\section{Ablations}

We use subjects A, B, and C for our ablation study. We consider the monocular setup described in \cref{sec:exp}. In addition to the qualitative results displayed in \cref{tab:ablations}, we also show the results of our ablation study qualitatively. \Cref{fig:ablation_qual} shows the effect of training our prior on differing numbers of subjects, ranging from using no prior, to using the complete 1k subjects. In each case, we select all frames from the first N training subjects in the synthetic training dataset for a prior with N subjects. \Cref{fig:ablation_qual} also shows the effect of using a different number of Gaussian primitives in the model. Here, we use varying UV map resolutions for the initialization (see \cref{sec:init}); we consider maps of resolution 64x64 (2926 Gaussians), 128x128 (11,758 Gaussians), 256x256 (46,928 Gaussians) and our full model using 512x512 (187,776 Gaussians).

\begin{figure}
    \centering
    \includegraphics[width=\columnwidth]{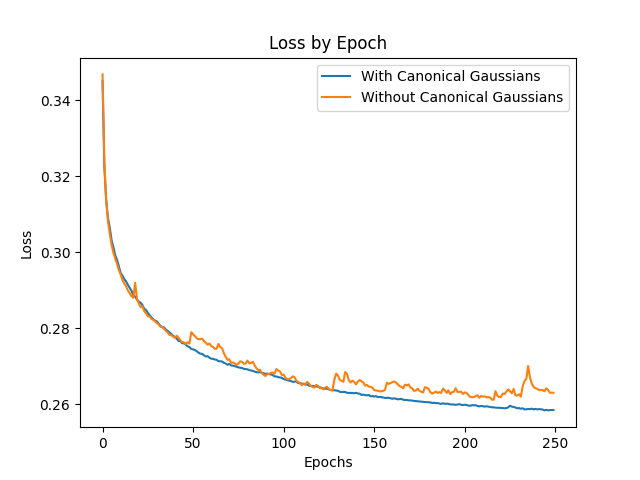}
    \caption{The training loss curves for $\lambda_{\textit{pix}}L_{\textit{pix}} + \lambda_{\textit{percep}}L_{\textit{percep}}$ with (blue) and without (orange) the canonical Gaussians. Note the improved training stability and better overall loss.}
    \label{fig:canon}
\end{figure}

\emph{Canonical Gaussians:} In addition to the ablations shown in the main paper, we also validate our claim that canonical Gaussians improve training stability. To show this, we plot the image space loss curves for $\lambda_{\textit{pix}}L_{\textit{pix}} + \lambda_{\textit{percep}}L_{\textit{percep}}$ in \cref{fig:canon}. 

\section{Ethical Concerns}

We recognise the potential for misuse of our model. We feel strongly about preventing this. We are actively researching watermarking methods for avatars and metadata labelling methods, such as the C2PA Initiative. We are also considering systems for likeness management, for example, only allowing a single account to operate an avatar. Before deploying any avatar system using our method, we will consult a wide range of stakeholders to mitigate the possibility of harm through our model.

Our model has advantages over others that have built priors over non-synthetic data. If we expose our prior to a user training their avatar, we do not run the risk of dataset distillation attacks. This means that there is no risk of privacy violations wherein an adversary could obtain personal data about subjects that have been used to train the prior. This also helps avoid legal issues around GDPR and consent. There is no chance of a subject withdrawing consent and requiring our prior to be retrained or detained.

\section{Comparison to Cafca}

\begin{figure}
    \centering
    \includegraphics[width=\columnwidth]{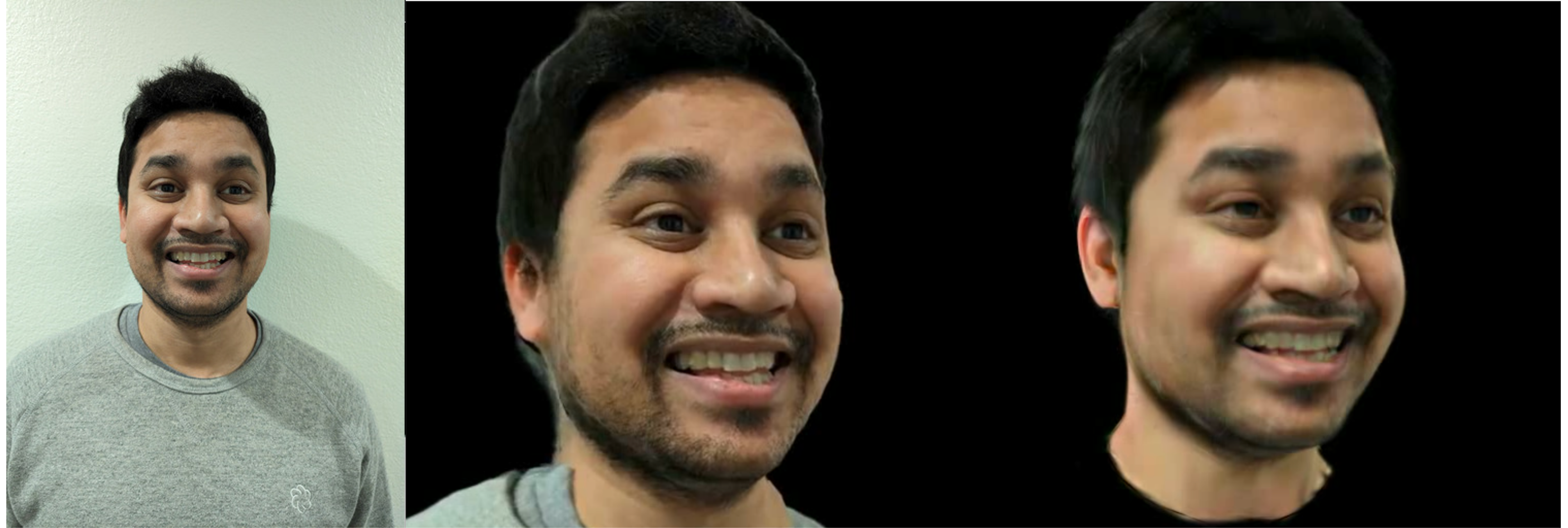}
    \caption{A comparison of our method (Right) compared to Cafca \cite{buehler2024cafca} (Middle), using the input image on the left. Our model performs better on the side of the head, such as on the ear, while being thousands of times faster to render. Our model can also be animated, while Cafca cannot.}
    \label{fig:cafca}
\end{figure}

Cafca \cite{buehler2024cafca} is a NeRF-based synthetic prior model that shares several similarities with our work. However, there is some crucial differences. Their method is only capable of modelling static expressions and cannot be animated. Furthermore, rendering for Cafca takes 20 seconds per frame on a 4 TPU machine. Our model, conversely can be freely animated and rendered at 70fps on a much more available NVIDIA 4090 RTX GPU. Despite our models much faster rendering time, we are able to achieve a similar level of quality, with our model better capturing the ear and back of head detail, but not quite getting as much high-frequency details. A comparison can be seen in \cref{fig:cafca}. As Cafca is not publically available, we take their results directly from their project page.

\end{document}